\documentclass[10pt,journal,compsoc]{IEEEtran}
\usepackage{graphicx}
\usepackage{array}
\usepackage{color}
\usepackage[dvipsnames]{xcolor}
\usepackage{times}
\usepackage[caption=false,font=footnotesize,labelfont=sf,textfont=sf]{subfig}
\usepackage{amsmath}
\usepackage{amssymb}
\usepackage{eqnarray}
\usepackage{algorithm}
\usepackage{algorithmicx}
\usepackage{algpseudocode}
\usepackage{xcolor}
\usepackage{tabularx}
\usepackage{colortbl}
\usepackage{epstopdf}
\usepackage{multirow}

\graphicspath{{./Figure/}}
\ifCLASSOPTIONcompsoc
  \usepackage[nocompress]{cite}
\else
  \usepackage{cite}
\fi

\begin{document}

\title{Unsupervised Embedding Learning from Uncertainty Momentum Modeling}

\author{Jiahuan~Zhou, Yansong Tang, Bing Su, Ying~Wu,~\IEEEmembership{Fellow,~IEEE}
\IEEEcompsocitemizethanks{\IEEEcompsocthanksitem Jiahuan Zhou and Ying Wu are with the Department of Electrical and Computer Engineering, Northwestern University, Evanston, IL, 60208. E-mail: zhoujh09@gmail.com, yingwu@northwestern.edu.\protect\\
\IEEEcompsocthanksitem Yansong Tang is with the Department of Engineering Science, University of Oxford, OX1 3PJ, UK, E-mail: tangyansong15@gmail.com.\protect\\
\IEEEcompsocthanksitem Bing Su is with the Beijing Key Laboratory of Big Data Management and Analysis Methods, Gaoling School of Artificial Intelligence, Renmin University of China, Beijing 100872, China. E-mail: subingats@gmail.com.\protect\\}
\thanks{Manuscript received xxxx.}}

\markboth{IEEE TRANSACTIONS ON PATTERN ANALYSIS AND MACHINE INTELLIGENCE}%
{Zhou \MakeLowercase{\textit{et al.}}: Unsupervised Embedding Learning from Uncertainty Momentum Modeling}

\IEEEtitleabstractindextext{%
\begin{abstract}

	Existing popular unsupervised embedding learning methods focus on enhancing the instance-level local discrimination of the given unlabeled images by exploring various negative data. However, the existed sample outliers which exhibit large intra-class divergences or small inter-class variations severely limit their learning performance. We justify that the performance limitation is caused by the gradient vanishing on these sample outliers. Moreover, the shortage of positive data and disregard for global discrimination consideration also pose critical issues for unsupervised learning but are always ignored by existing methods. To handle these issues, we propose a novel solution to explicitly model and directly explore the uncertainty of the given unlabeled learning samples. Instead of learning a deterministic feature point for each sample in the embedding space, we propose to represent a sample by a stochastic Gaussian with the mean vector depicting its space localization and covariance vector representing the sample uncertainty. We leverage such uncertainty modeling as momentum to the learning which is helpful to tackle the outliers. Furthermore, abundant positive candidates can be readily drawn from the learned instance-specific distributions which are further adopted to mitigate the aforementioned issues. Thorough rationale analyses and extensive experiments are presented to verify our superiority.
\end{abstract}

\begin{IEEEkeywords}
	unsupervised embedding learning, uncertainty modeling, distribution learning, vanishing gradient.
\end{IEEEkeywords}
}

\maketitle

\IEEEdisplaynontitleabstractindextext

\IEEEpeerreviewmaketitle

\newtheorem{theorem}{Theorem}
\newtheorem{proof}{Proof}
\section{Introduction}
\label{sec:intro}
As an important and long-standing problem in computer vision, learning effective and versatile image feature representations for downstream computer vision tasks has been consistently attractive. During the past decades, supervised feature embedding learning~\cite{russakovsky2015imagenet} has achieved promising performance whose success heavily relies on a massive collection of accurately labeled training data. However, collecting such large-scale annotated images for learning is not only time-consuming but also costly. Therefore, in recent years, unsupervised feature embedding learning (also known as self-supervised feature learning)~\cite{dosovitskiy2015discriminative,wu2018unsupervised,ye2019unsupervised,he2020momentum,chen2020simple,grill2020bootstrap,caron2020unsupervised,hu2021adco} has attracted increasing attention in various fundamental research areas~\cite{su2017unsupervised,zhou2017efficient,deng2009imagenet}.

\begin{figure}[]
	\begin{center}
		\includegraphics[width=0.5\textwidth]{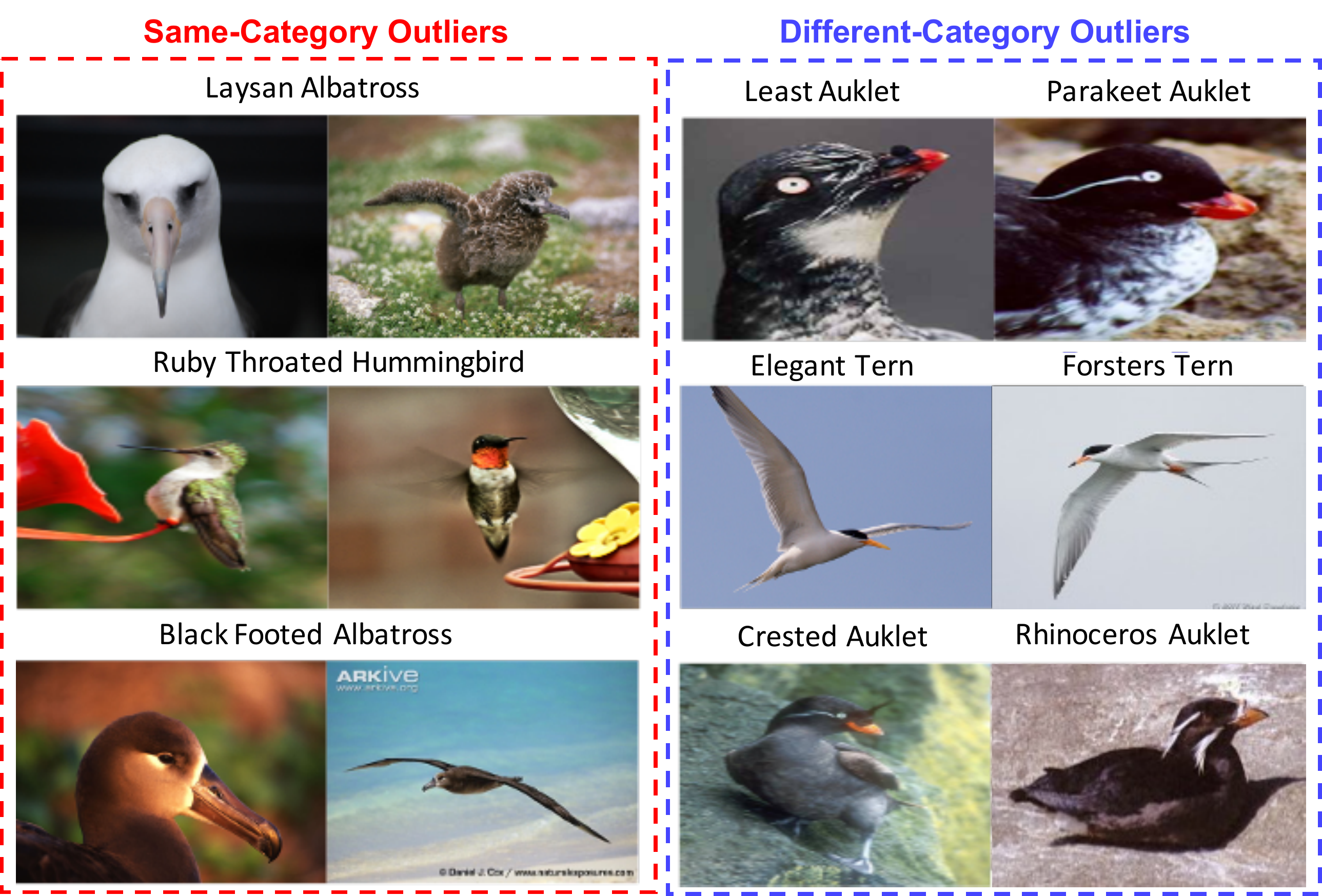}
	\end{center}
	\caption{The critical unlabeled sample outliers are the main source of data noises for unsupervised embedding learning. Several cases in CUB200~\cite{wah2011caltech} are presented (various bird categories are shown). \textcolor{red}{\textbf{Left}} illustrates even two instances are from the same category, the viewpoint changes, instance characteristics difference or occlusions may cause outliers. \textcolor{blue}{\textbf{Right}} shows even two instance are from different categories, they could be closely visually similar to each other. (Best view in color)}
	\label{F:issue}
\end{figure}

\begin{figure*}[]
	\begin{center}
		\includegraphics[width=1\textwidth]{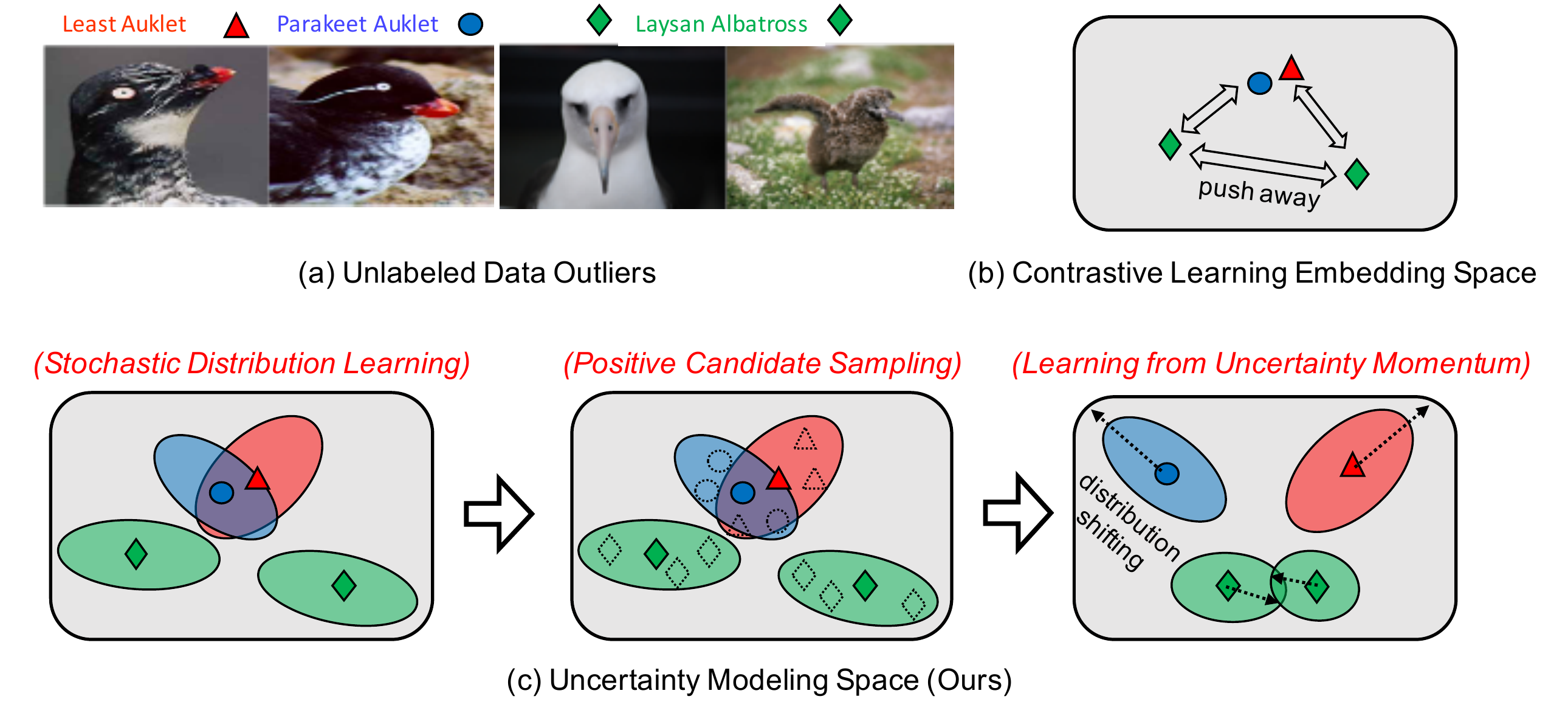}
	\end{center}
	\caption{\label{F:motivation} For a learning minibatch, the presented unlabeled data outliers in (a) always exist. The performance of typical unsupervised embedding learning methods is severely limited by these outliers with large intra-class variations or small inter-class divergences. It is because they fail to push away the highly similar samples from different categories and sacrifice the feature compactness of the same-category samples as in (b). Instead, by well modeling and exploring the uncertainty of unlabeled data, our method can gradually enhance the discriminative ability of these outliers in the obtained uncertainty modeling space by sampling abundant positive candidates for learning. The details of our idea are introduced in Sec.~\ref{sec:tech}. (Best view in color)}	
\end{figure*}

To alleviate the requirement of large-scale label annotations for supervised approaches, existing unsupervised feature embedding learning methods~\cite{dosovitskiy2015discriminative,wu2018unsupervised,ye2019unsupervised,he2020momentum,chen2020simple,grill2020bootstrap,caron2020unsupervised,hu2021adco} typically regards each image as an individual instance in learning to enforce them separable to each other as much as possible. Their focus is concentrated on how to enhance the instance discrimination by exploring various negative samples that are stored by a memory bank~\cite{wu2018unsupervised}, maintained by a dynamic queue with a momentum update~\cite{he2020momentum}, or represented by the other samples in the same minibatch~\cite{chen2020simple}. To do so, an instance-to-instance contrastive learning strategy is adopted and such instance-level local discrimination enhancement makes these methods fairly scalable to the given unlabeled images and has achieved promising performance.

In practice, their learning performance is largely influenced by several critical issues including (1) the unlabeled sample outliers, (2) extremely limited positive data, and (3) the disregard for global discrimination consideration. In fact, the aforementioned unsupervised solutions take the assumption that all the given unlabeled images are reliable and of good quality for granted. However, due to no label information is available, the given data may severely suffer from sample outliers that are the main source of data noises in unsupervised learning. As illustrated in Fig.~\ref{F:issue}, for the instances which are from different categories but highly visually similar to each other, the aforementioned methods fail to separate them based on their adopted instance-wise contrastive learning strategy because of the vanishing gradient problem (more detailed theoretical analyses are discussed in Sec.~\ref{sec:theorem}). Other instances that are factually from the same category are visually outlying because of viewpoint changes, instance characteristics differences, occlusions, and other factors. Existing unsupervised embedding learning methods still follow the same principle to maximally push them farther away from each other. Although the separability of learned features is enhanced, they severely sacrifice the feature compactness of the same-category samples. For unsupervised embedding learning, such sample outliers always exist and directly learning from these noisy samples has to sacrifice the inter-class separability and intra-class compactness which leads to performance degradation. Therefore, the treatment regardless of sample outliers is the main root to limit the performance of existing unsupervised embedding learning approaches. 

Meanwhile, the shortage of positives for learning is another important issue in unsupervised embedding learning which degrades the performance of existing approaches. Since each unlabeled instance itself is the only positive sample for learning, existing solutions to alleviate this issue are leveraging various data augmentation techniques~\cite{chen2020simple} but the quality of augmented data is hard to control. On the other hand, because it is hard to obtain useful positive samples for learning, existing instance-level local discrimination enhancement relies on the instance-to-instance contrastive learning against the negatives. Thus, the global separability and compactness of an instance over all the other instances may not be fully optimized since no global discrimination enhancement is considered within a learning minibatch.

To tackle the above crucial issues, we propose a novel solution that aims to model and explore the uncertainty of the given unlabeled samples for learning. As illustrated by Fig.~\ref{F:motivation}, instead of learning a deterministic feature point for each instance in the embedding space, we propose to represent a sample by a multivariate Gaussian with the mean vector depicting the localization in embedding space and covariance vector representing the sample uncertainty. Such an idea is inspired by the research of uncertainty modeling and distribution learning~\cite{geng2013facial,tzelepis2017linear,oh2018modeling}. Unlike some works which focus on how to diminish the influence of uncertainty for better learning, we are devoted to introducing feature uncertainty as momentum to the learning which is verified to be helpful to tackle the outliers by our experiment results. Another appealing merit of our method is abundant positive candidates can be readily drawn from the learned instance-specific distribution, so the limited positives in unsupervised feature embedding learning can be mitigated. We propose a novel set-to-set similarity-based loss to alleviate the unstable nor ineffective loss optimization by evaluating only a point-to-point similarity as previous methods do. Moreover, to further enhance the discrimination of learned embedding space and stabilize the learning process, a distribution consistency loss and a minibatch-wise global ranking loss are proposed to integrate with the above classification loss in our learning. Thorough theoretical justifications and analyses are discussed to demonstrate the necessity of sample uncertainty modeling. Extensive experiments on various evaluation benchmarks (ImageNet~\cite{russakovsky2015imagenet}, CIFAR-10~\cite{krizhevsky2009learning}, STL-10~\cite{coates2011analysis}, CUB200-2011 (\textit{CUB200})~\cite{wah2011caltech} Car196~\cite{krause20133d} and Stanford Online Product (\textit{Product})~\cite{oh2016deep}) verify the superiority and effectiveness of our proposed model against the state-of-the-art unsupervised feature embedding learning approaches.
\section{Related Work}
\label{sec:work}
\subsection{Contrastive Unsupervised Embedding Learning}
In recent years, many unsupervised embedding learning methods rely on contrastive learning by treating each sample itself as an individual class to maximally separate them in the learned embedding space. Among them, most works focus on enhancing the instance-wise local discrimination by exploring various negative samples that are stored by a memory bank~\cite{wu2018unsupervised}, maintained by a dynamic queue with a momentum update~\cite{he2020momentum,hu2021adco}, or represented by the other samples in the same minibatch~\cite{chen2020simple}. To mitigate the influence of tremendous negative-pair comparisons, \cite{grill2020bootstrap} adopted a set of trainable prototype vectors to represent all the samples. Besides, some variants of contrastive learning regularizations are proposed to enhance embedding learning. \cite{zhuang2019local} proposed to optimize a metric of local aggregation to enhance the discrimination of unlabeled learning samples. \cite{ye2019unsupervised} aimed at learning data augmentation invariant and instance spread-out features for the given unlabeled samples by directly optimizing the ``real'' instance features on top of the Softmax function. \cite{ye2020probabilistic} proposed to learn a probabilistic structural latent representation for unsupervised samples which approximates the positive concentrated and negative instance separated properties in the graph latent space.

However, all of the aforementioned methods aim to learn a deterministic feature vector representation for the given unlabeled samples in the obtained embedding space. Since no specific sample uncertainty modeling is considered, their overall learning performance is largely limited if the unlabeled data contain severe sample outliers (e.g., highly similar samples but of different categories as shown in Fig.~\ref{F:issue}). Instead, our proposed method manages to model and leverage the uncertainty of unlabeled samples by learning an instance-specific distribution for each datum hence the outlier issue can be greatly mitigated.

\subsection{Clustering Unsupervised Embedding Learning}
Besides the above contrastive learning-based approaches, by appropriately assigning pseudo-labels to the learning samples, clustering-based unsupervised embedding learning methods also achieve promising performance.
\cite{dosovitskiy2015discriminative} proposed a discriminative objective by generating a set of surrogate labels via invariant data augmentation. \cite{caron2018deep} designed a scalable clustering approach for the unsupervised representation learning by iterating between clustering with k-means the learned features and updating its weights for pseudo-label assignment in a discriminative loss. Meanwhile, \cite{caron2020unsupervised} explored a strategy of iterative clustering and pseudo-label generation to alleviate the pair-wise comparisons among learning samples. \cite{li2020prototypical} introduced prototypes as latent variables and performed iterative clustering and representation learning in an EM-based framework to implicitly encode semantic structures of the data in the learned embedding space. However, for these mentioned clustering-based methods, the correctness and quality of generated pseudo-labels are hard to guarantee so their learning performance is limited. Moreover, the costly clustering processing also limits their practical usage.

\begin{figure*}[]
	\begin{center}
		\includegraphics[width=1\textwidth]{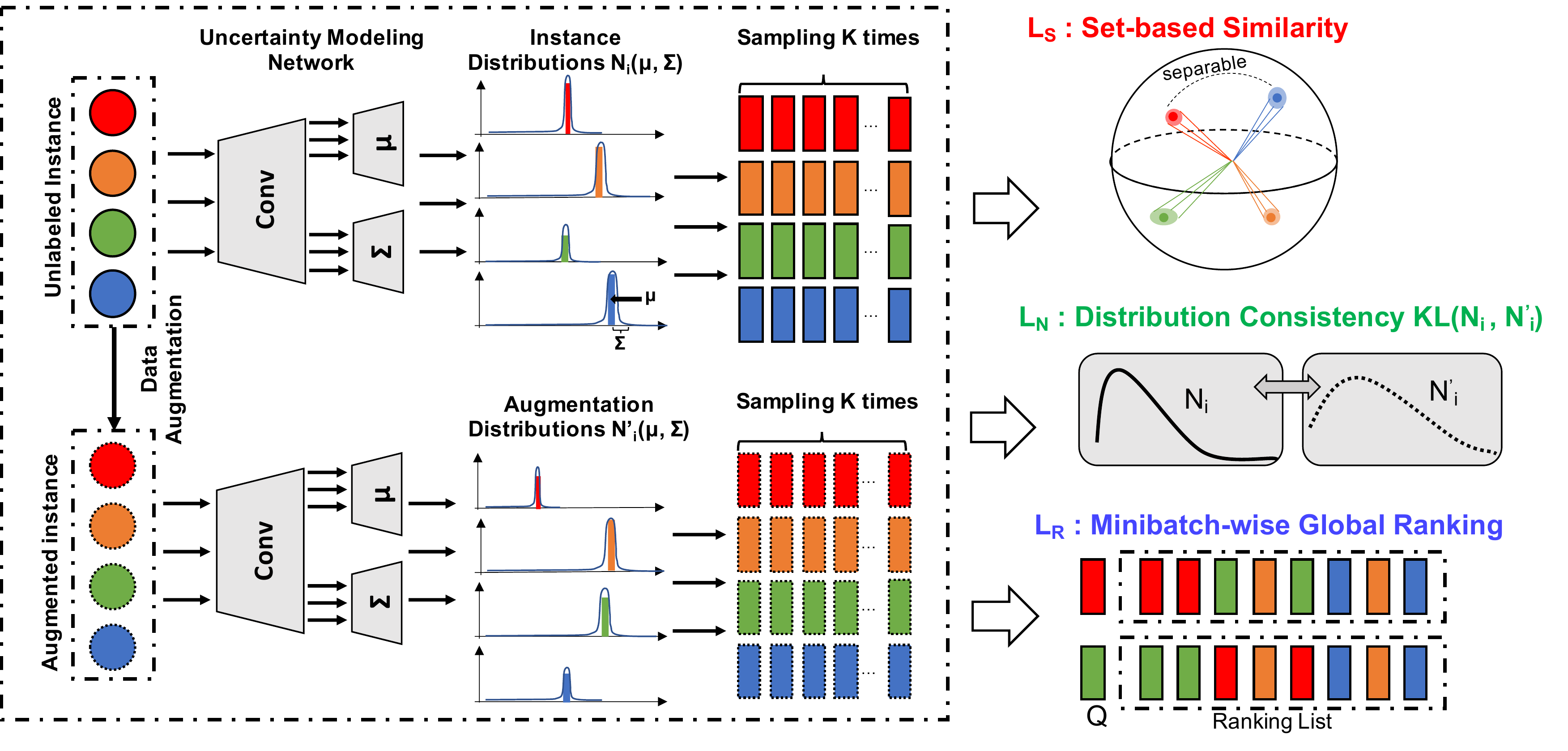}
	\end{center}
	\caption{\label{F:pipeline}The pipeline of our proposed unsupervised embedding learning from uncertainty momentum modeling. The given unlabeled images are fed into the uncertainty modeling network that each instance $x_i$ will be explicitly modeled as a multivariate Gaussian distribution with mean vector $\mu_i$ and diagonal covariance vector $\Sigma_i$. Then multiple positive candidates $\{z_k^i\}_k$ of instance $x_i$ will be randomly drawn from the learned distribution to form a reliable set-based representation for $x_i$ which are utilized to compute the set-to-set distance similarity for discriminative loss learning.}
\end{figure*}

\subsection{Deep Uncertainty Modeling and Distribution Learning}
Our idea is initially motivated by existing researches about uncertainty modeling in DNN which has attracted more and more attention in various computer vision tasks. \cite{oh2018modeling} proposed to utilize the variational information bottleneck principle for category-level uncertainty modeling to facilitate image classification. For object detection, \cite{choi2019gaussian} proposed a method for predicting the localization uncertainty that indicates the reliability of bbox aiming at reducing the false positive and increasing the true positive for improving the detection accuracy. \cite{he2019bounding} proposed to tackle the ambiguities introduced by labeling the bounding boxes via a novel bounding box regression loss for learning bounding box transformation and localization variance together. For medical image processing, \cite{zhang2019reducing} aimed to model the reconstruction uncertainty in Magnetic Resonance Imaging (MRI) reconstruction by dynamically selecting the measurements to take and iteratively refine the prediction during the inference phase. \cite{patro2019u} provided a method that uses gradient-based certainty attention regions to obtain better visual question answering. The obtained improving uncertainty estimates correlate better with misclassified samples. \cite{yu2019robust} focused on learning deep person re-identification (P-RID) models that are robust against label noises of training data by modeling the uncertainty of the extracted features. \cite{bertoni2019monoloco} addressed the ambiguity in monocular 3D pedestrian localization by predicting confidence intervals through a loss function based on the Laplace distribution. \cite{subedar2019uncertainty} proposed an uncertainty-aware multi-modal Bayesian fusion framework for audiovisual activity recognition which focuses on well handling the predictive uncertainty associated with individual modalities. Compared to these methods, we propose to model and directly utilize the instance-level uncertainty of unlabeled samples themselves which is more challenging than the above ones that focus on diminishing the uncertainty of supervision labels. In addition, the aforementioned methods cannot handle the uncertainty of the given data if the label information is not provided, but our method does.

In addition, several recent works attempt to represent the feature of images as a distribution. This is inspired by the pioneering work in facial age estimation~\cite{geng2013facial} that first proposed a label distribution learning (LDL) model for facial age estimation by assigning an age distribution to each face image. Recently, \cite{tzelepis2017linear} aimed to model each example as a multi-dimensional Gaussian distribution described by its mean vector and covariance matrix. \cite{wan2018rethinking} proposed a large-margin Gaussian Mixture (LGM) loss for image classification under the assumption that the deep features of the learning data follow a Gaussian Mixture distribution. However, these feature distribution learning approaches heavily rely on large-scale labeled data as the supervision of uncertainty modeling and the generalization ability of the learned category-level distribution is limited to unseen testing samples since no instance-specific uncertainty adaptation is in the loop.
\section{Unsupervised Embedding Learning from Uncertainty Momentum Modeling}
\label{sec:tech}
The main objective of unsupervised deep embedding learning is to project the given unlabeled images $\mathcal{I} = \{x_1, x_2, \dots, x_n\}$ in a minibatch to a $D$-dimensional discriminative feature embedding space $\mathcal{V} = \{v_1, v_2, \dots, v_n\}$ via the learned deep neural network $f_{\theta}$:
\begin{equation}
	v_i = f_{\theta}(x_i).
\end{equation}

Since no supervision information is provided for $\mathcal{I}$, each instance in $\mathcal{I}$ is considered as an individual category, and the learning of $f_{\theta}$ aims to make them distinguishable to each other as much as possible in the learned embedding space.

\subsection{From \textit{Deterministic Point Learning} to \textit{Stochastic Distribution Estimation}}
For each $x_i \in \mathcal{I}$, instead of learning a deterministic feature representation vector in the embedding space which is fragile to the data outliers~\cite{dosovitskiy2015discriminative,wu2018unsupervised,ye2019unsupervised,he2020momentum,chen2020simple}, we aim to model the uncertainty of $v_i$ produced by $f_{\theta}$ via learning a multivariate Gaussian over the vector that is parameterized by a $D$-dimensional mean vector $\mu$ and a $D$-dimensional diagonal covariance matrix $\Sigma$:
\begin{equation}
	p(v_i|x_i) = \mathcal{N}(v_i|\mu(x_i), \Sigma(x_i)),
	\label{eqn:distribution}
\end{equation}
where $\mu(x_i)$ aims at depicting the spatial localization of $x_i$ in the embedding space and $\Sigma(x_i)$ models the uncertainty of $v_i$. From the probability perspective, the feature vector $v_i$ of image $x_i$ is considered as a random variable drawn from the above distribution. To achieve this and enhance computational efficiency, two separate feature extraction layers are added to the backbone network for predicting $\mu(x_i)$ and $\Sigma(x_i)$ respectively. Besides, it is worth mentioning that instead of modeling a $D \times D$ full covariance matrix which is computationally expensive, we consider $\Sigma(x_i)$ to be diagonal which reduces its dimension from $D^2$ to $D$.

\textbf{Extend to Mixture of Gaussians (MoG)}:
Considering that real-word images usually exhibit complex distributions caused by appearance variations, a single Gaussian may not be sufficient to fully represent a sample. Therefore, we propose to extend the uncertainty modeling as a mixture of Gaussians (MoG). To do so, $C$ independent concatenated branches consisted of two separate feature extraction layers are added to the backbone network for predicting all $\mu(x_i,c)$ and $\Sigma(x_i,c)$ respectively.
\begin{equation}
	p(v_i|x_i) = \sum_{c=1}^{C}\mathcal{N}(v_i|\mu(x_i,c), \Sigma(x_i,c)).
	\label{eqn:distribution-2}
\end{equation}

Similarly, each $\mu(x_i,c)$ aims at depicting the spatial localization of a sub-distribution in the embedding space and $\Sigma(x_i,c)$ models the corresponding sub-distribution uncertainty. Such MoG uncertainty modeling can readily tackle the multi-modal distribution of practical images. For the sake of convenience, in the following descriptions,
we use the single Gaussian modeling in Eqn.~\ref{eqn:distribution} to demonstrate our method and adopt $\mathcal{N}_i$, $\mu_i$, and $\Sigma_i$ as the abbreviations for $\mathcal{N}(v_i|\mu(x_i), \Sigma(x_i))$, $\mu(x_i)$, and $\Sigma(x_i)$ respectively.

\subsection{Positive Extraction via Monte-Carlo Sampling}
\label{sec:sampling}
A crucial merit introduced by modeling the sample uncertainty via Eqn.~\ref{eqn:distribution} or Eqn.~\ref{eqn:distribution-2} is that multiple positive feature candidates of $x_i$ can be readily obtained by performing random Monte-Carlo sampling from the learned distributions to form an informative positive candidate set $\mathcal{Z}_i = \{z_i^1, z_i^2, \dots, z_i^k\}$ for instance $x_i$ (all $z_i^j$ are $l_2$-normalized).
\begin{equation}
	p(v_i|x_i) \xrightarrow{\text{sampling}} \mathcal{Z}_i,
\end{equation}
where the sampled positive candidates in $\mathcal{Z}_i$ can be considered as a set of independent and identically distributed (I.I.D) random variables. Compared with previous solutions~\cite{dosovitskiy2015discriminative,wu2018unsupervised,ye2019unsupervised,he2020momentum,chen2020simple} that only estimate a single feature point, $\mathcal{Z}_i$ provides more diverse information to depict the local discrimination of $x_i$ that is helpful and crucial to tackle sample outliers in $\mathcal{I}$ as analyzed in Sec.~\ref{sec:theorem}. In addition, $\mathcal{Z}_i$ can be directly explored as the positive data-augmentation for $x_i$ as well as the negative augmentation for the other instances.

However, sampling such a positive feature candidate set $\mathcal{Z}_i$ for $x_i$ introduces more memory storage burden and computation cost than only keeping one single feature point. To alleviate this, we propose a dynamic sampling-from-distribution (SFD) strategy by considering the distribution learning process. In the early stage of distribution learning, the uncertainty of $x_i$ is large so that more feature candidates in $\mathcal{Z}_i$ should be drawn to cover the variance. As the learning progresses, instances in $\mathcal{I}$ become more and more discriminative and the uncertainty of instance $x_i$ becomes smaller, hence fewer feature candidates are enough for distribution representation. The effectiveness and rationale of adopting SFD during distribution learning have been verified by the experiments in Sec.~\ref{sec:exp}.

\subsection{Uncertainty-Aware Global-Local Instance Discrimination Enhancement}
To enhance the discrimination of learned representations, existing contrastive learning-based unsupervised embedding learning methods rely on instance-wise similarity measurement. No matter optimizing the discrimination of given samples with the classifier weights~\cite{dosovitskiy2015discriminative}, the memory bank~\cite{wu2018unsupervised,he2020momentum} or the invariant data augmentations~\cite{ye2019unsupervised,chen2020simple}, the instance-wise similarity is restricted to a pair-wise point-to-point distance comparison. However, due to the absence of label information, such a point-to-point distance is sensitive to the sample outliers (e.g., near-identical instances from different classes) which results in unstable and unreliable learning performance.

\textbf{Set-to-Set Softmax Loss $\mathcal{L}_{S}$}: owing to the sampled positive candidate set $\mathcal{Z}_i$ for instance $x_i$, we can readily optimize a set-to-set similarity measurement between $\mathcal{Z}_i$ and $\mathcal{Z}_j$ so that the inter-instance separability and intra-instance compactness of $x_i$ in the learned embedding space can be jointly enhanced. To do so, we adopt a simple but effective \textit{Mean Average Inner Product} distance as the set-to-set similarity metric in Eqn.~\ref{eqn:set-to-set distance}~\footnote{Various options such as hull-based distance~\cite{zhu2013point}, earth-mover distance~\cite{rubner2000earth} can be easily utilized here.}.
\begin{equation}
	d(x_i,x_j) = \dfrac{1}{k^2}\sum_{u=1}^k\sum_{t=1}^k(z_u^i)^Tz_t^j.
	\label{eqn:set-to-set distance}
\end{equation}

Therefore, by replacing the point-to-point inner product distance with our Eqn.~\ref{eqn:set-to-set distance}, a set-to-set similarity driven Softmax optimization loss can be readily obtained as follows. Similar to \cite{ye2019unsupervised,chen2020simple,he2020momentum}, for an instance $x_i$ (its feature candidate set is $\mathcal{Z}_i$), its corresponding invariant data augmentation counterpart is $x'_i$. We can obtain the feature candidate set $\mathcal{Z}'_i$ for $x'_i$ following the same operation to $x_i$. So the probability of $x_i$ is correctly recognized as the same category as $x'_i$ is:
\begin{equation}
	P(x'_i|x_i) = \frac{\exp(d(x_i,x'_i)/\tau)}{\sum_{k=1}^n\exp(d(x_k,x'_i)/\tau)},
	\label{eqn:softmax-same}
\end{equation}
where $\tau$ is a temperature parameter as in \cite{wu2018unsupervised}. The final Softmax classification loss is modeled as minimizing the sum of negative log likelihood over all samples:
\begin{equation}
	\label{eqn:ls}
	\mathcal{L}_{S} = -\sum_{i=1}^n\log P(x'_i|x_i).
\end{equation}

As demonstrated by Eqn.~\ref{eqn:ls}, a smaller loss is obtained if the sampled positives in $\mathcal{Z}_i$ and $\mathcal{Z}'_i$ are more compact as well as the candidates in $\mathcal{Z}_i$ and $\mathcal{Z}_j$ are more separable to each other.

\begin{figure*}[]
	\begin{center}
		\includegraphics[width=1\textwidth]{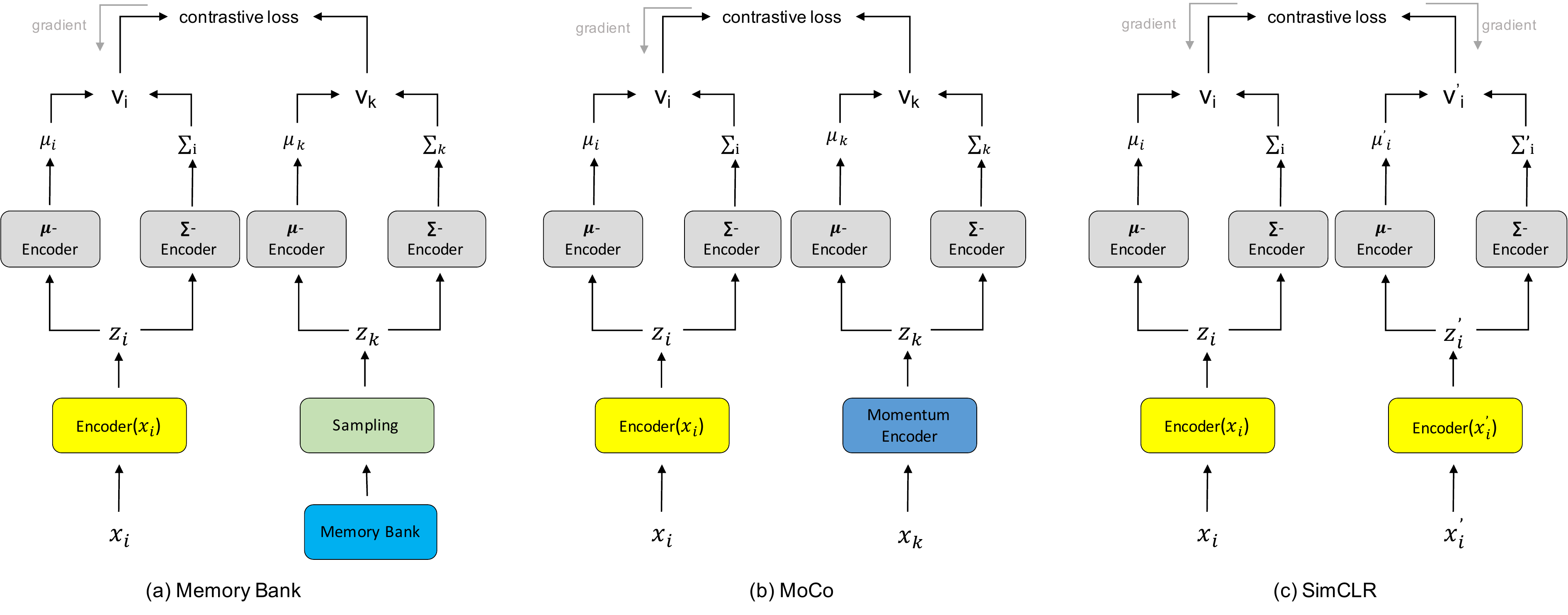}	
	\end{center}
	\caption{\label{F:models} For most of the existing mainstream unsupervised embedding learning networks, our proposed uncertainty momentum model can be directly and readily integrated into them by only modifying the feature output layer of the backbone network.}
\end{figure*}

\textbf{Distribution Consistency Loss $\mathcal{L}_{\mathcal{N}}$}: while solely optimizing the above Softmax classification loss $\mathcal{L}_{S}$ is not sufficient to drive the sampled feature candidates towards the learned Gaussian distribution directly because a sampled positive $z_u^i$ for instance $x_i$ can be far away from the corresponding distribution mean centroid $\mu_i$ while still being correctly classified as long as it is relatively closer to $\mu_i$ than to the other means of the other instances. Therefore, to explicitly enhance the consistency of learned distributions, we propose to minimize the distribution divergence over all $\{x_i, x'_i\}$ pairs. The Kullback-Leibler (KL) divergence between two instance-specific multivariate Gaussian distributions $\mathcal{N}_i$ and $\mathcal{N}_j$ can be measured as:
\begin{equation}
	\begin{aligned}
		&D_{KL}(\mathcal{N}_i||\mathcal{N}_j) = \dfrac{1}{2}(\log\dfrac{|\Sigma_j|}{|\Sigma_i|}-D+ \\
		&tr\{\Sigma_j^{-1}\Sigma_i\}+(\mu_j-\mu_i)^T\Sigma_j^{-1}(\mu_j-\mu_i)).\\
	\end{aligned}
\end{equation}

Inspired by the Jensen–Shannon (JS) divergence metric which is symmetric, finite and non-negative, the distribution consistency between $x_i$ and $x'_i$ can be measured as:
\begin{equation}
	D(x_i,x'_i) = D_{KL}\left(\mathcal{N}_i||\mathcal{N}'_i\right) + D_{KL}\left(\mathcal{N}'_i||\mathcal{N}_i\right).
\end{equation}

Therefore, the proposed distribution consistency loss is to minimize $\mathcal{L}_{\mathcal{N}}$ as in Eqn.~\ref{eqn:L_N}:
\begin{equation}
	\mathcal{L}_{\mathcal{N}} = \sum_{i=1}^n D(x_i,x'_i).
	\label{eqn:L_N}
\end{equation}

\textbf{Minibatch-wise Global Ranking Loss $\mathcal{L}_{R}$}: both the above set-to-set Softmax loss $\mathcal{L}_{S}$ and distribution consistency loss $\mathcal{L}_{\mathcal{N}}$ are still restricted by the similarity between only two instances $x_i$ and $x_j$ in a minibatch. However, for a feature candidate of $x_i$ in the minibatch, it should be more similar to all of its same-category candidates than the other samples. Such a kind of global similarity relationship within the whole minibatch should be seriously considered but always ignored previously. Therefore, we propose a minibatch-wise global ranking loss to enhance the global intra-class compactness information. Considering a feature candidate $z_i^j$ of $x_i$ in the minibatch as a query, the ranking list of all the samples in all candidate sets $\{\mathcal{Z}_i\}_{i=1}^n$ can be obtained as $\mathcal{R}_{z_i^j} = (z_1,z_2, ..., z_{nk})$ based on the cosine similarity. Each $z_K$ in $\mathcal{R}_{z_i^j}$ is a feature candidate in the collection of all $\{\mathcal{Z}_i\}_{i=1}^n$. Thus the average precision (AP) of $z_i^j$ is the average of precision values ($P_i^j@K$) evaluated at different positions:
\begin{equation}
	AP_i^j = \dfrac{1}{k}\sum_{K=1}^{nk}\textbf{1}[z_K \in \mathcal{Z}_i]P_i^j@K,
\end{equation}
where $\textbf{1}[.]$ is the binary indicator and $P_i^j@K = \dfrac{1}{K}\sum_{u=1}^{K}\textbf{1}[z_u \in \mathcal{Z}_i]$ will achieve the optimal value if and only if all candidates in $\mathcal{Z}_i$ are ranked above all the other samples. Finally, the minibatch-wise global ranking loss $\mathcal{L}_{R}$ is modeled as the difference between 1 and the mean average precision (mAP) over all the obtained candidates:

\begin{equation}
	\begin{aligned}
		\mathcal{L}_{R} & = 1-\dfrac{1}{nk}\sum_{i=1}^{n}\sum_{j=1}^{k}AP_i^j\\
		& = 1-\dfrac{1}{nk^2}\sum_{i=1}^{n}\sum_{j=1}^{k}\sum_{K=1}^{nk}\textbf{1}[z_K \in \mathcal{Z}_i]P_i^j@K \\
		& \approx 1-\dfrac{1}{nk^2}\sum_{i=1}^{n}\sum_{j=1}^{k}\sum_{z_K \in \mathcal{Z}_i}\sum_{b=0}^{B}\textbf{1}[d_h(z_K, z_i^j) = b] \\
		& \approx 1-\dfrac{1}{nk^2}\sum_{i=1}^{n}\sum_{j=1}^{k}\sum_{z_K \in \mathcal{Z}_i}\sum_{b=0}^{B}\delta(d_h(z_K, z_i^j),b), \\
	\end{aligned}
\end{equation}
where $d_h(z_K, z_i^j)$ is the discrete histogram quantization distance whose values are within a range of \{0, $B$\} based on the original continue cosine similarity distance as introduced by \cite{he2018local}. Therefore, the above approximation can be theoretically guaranteed by performing histogram quantization operation to the cosine similarity among candidates and replacing the binary indicator with a differentiable $\delta$ function~\cite{simo2015discriminative,he2018hashing}. Therefore, the $\mathcal{L}_R$ is differentiable that can be easily optimized via back-propagation.

Finally, the overall loss function $\mathcal{L}$ for our unsupervised embedding learning from uncertainty momentum modeling contains three components: the set-to-set distance-based Softmax loss $\mathcal{L}_{S}$, the distribution consistency loss $\mathcal{L}_{\mathcal{N}}$ and the minibatch-wise global ranking loss $\mathcal{L}_{\mathcal{R}}$:
\begin{equation}
	\label{eqn:final_loss}
	\mathcal{L} = \mathcal{L}_{S} + \lambda_\mathcal{N}\mathcal{L}_{\mathcal{N}} + \lambda_\mathcal{R}\mathcal{L}_{\mathcal{R}},
\end{equation}
where $\lambda_\mathcal{N}$ and $\lambda_\mathcal{R}$ are the weighting parameters for scale balance. Via minimizing $\mathcal{L}$, the network can be accordingly optimized as shown in Fig.~\ref{F:pipeline}.


\subsection{Connection to Existing Unsupervised Methods}
As demonstrated in Fig.~\ref{F:models}, our proposed uncertainty momentum modeling can be readily and directly adopted by most existing unsupervised embedding learning methods (Memory bank-based models~\cite{wu2018unsupervised}, MoCo-guided methods~\cite{he2020momentum,chen2020improved,hu2021adco}, end-to-end networks~\cite{chen2020simple,ye2019unsupervised,ye2020probabilistic}) which aim to learn a deterministic feature point for the given learning data:

\begin{figure*}[]
	\begin{center}
		\includegraphics[width=1\textwidth]{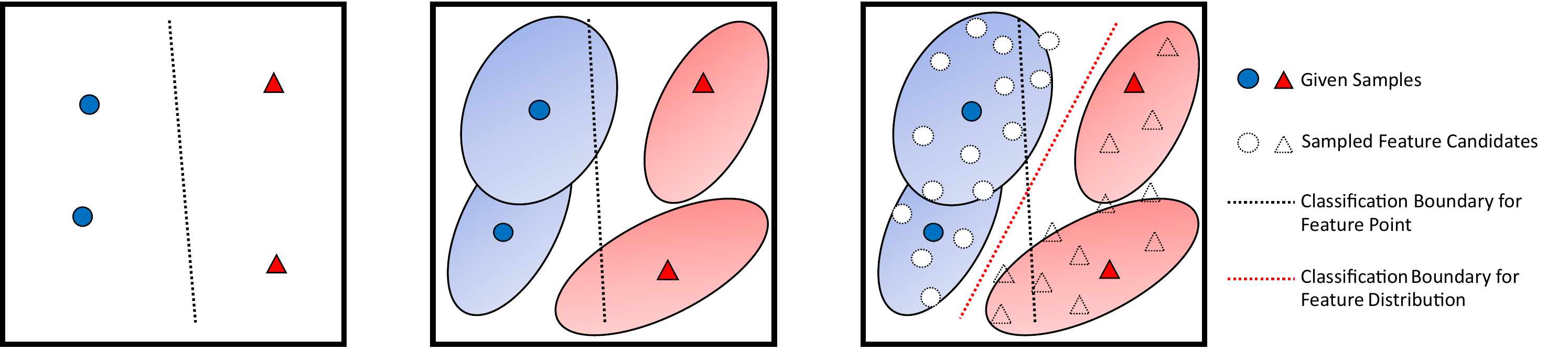}
	\end{center}
	\caption{\label{F:explain}The explanation of our proposed uncertainty momentum modeling via PAC-Bayesian theory. Compared with existing feature point learning-based methods, the classification boundary can be refined by our uncertainty modeling so that the generalization error bound can be further reduced.}
\end{figure*}

\begin{itemize}
	\item Memory bank-based methods~\cite{wu2018unsupervised}: the data encoder for $\{x_i\}_{i=1}^n$ and stored samples in the memory bank will be modeled via the proposed $\mu$-encoder and $\Sigma$-encoder for feature distribution learning. For the training, the feature candidates will be extracted as described in Sec.~\ref{sec:sampling} which will further be utilized for the proposed learning loss Eqn.~\ref{eqn:final_loss}. For the testing, the obtained mean vector from $\mu$-encoder will be directly used as the feature representation $v_i$ for the query sample $x_i$.
	\item MoCo-guided methods~\cite{he2020momentum,chen2020improved,hu2021adco}: the data encoder for $\{x_i\}_{i=1}^n$ and the momentum encoder for the queue will be modeled via the proposed $\mu$-encoder and $\Sigma$-encoder to for feature distribution learning. For the training and testing, the same process as described in \textit{Memory bank-based methods} will be adopted.
	\item End-to-end methods: the encoders for both the learning data $\{x_i\}_{i=1}^n$ and their augmented counterparts $\{x_i'\}_{i=1}^n$ will be modeled via our designed $\mu$-encoder and $\Sigma$-encoder for feature distribution learning. For the training and testing, the same process as described in \textit{Memory bank-based methods} will be adopted.
\end{itemize}

\subsection{Justifications and Analyses}
\label{sec:theorem}
\subsubsection{The Influence of Unlabeled Sample Outliers}
For a given training minibatch $\mathcal{I}$, since the images in $\mathcal{I}$ are completely unlabeled without knowing the labels and data distributions, the unlabeled sample outliers which are the main source of data noises will always exist and largely influence the embedding learning~\cite{yu2019robust}. Below, we will thoroughly analyze the outlier influence to the point-to-point similarity-based contrastive learning methods and our proposed set-to-set distance-based one under the scenario that two images $x_i$ and $x_j$ are located around the classification boundary which makes them almost identical to each other (More critically, they are the same images but redundantly collected in $\mathcal{I}$ by mistake, $x_i = x_j$). For the cosine distance-based Softmax loss Eqn.~\ref{eqn:softmax-diff-2} in existing methods, since the inner products $v_i^Tv_j$ and $v_j^Tv_j$ are 1 and $\tau$ is small, the value of the whole denominator is determined by $\exp(v_j^Tv_j/\tau)$. So $P(x_i|x_j) \approx 1$ always holds no matter what the other samples are. Even existing unsupervised methods treat each image as an individual category aiming at making them separable in the embedding space by optimizing Eqn.~\ref{eqn:softmax-diff-2}, they can not tackle these outlier noises at all since $P(x_i|x_j) \approx 1$ provides no helpful back-propagated gradients in learning.
\begin{equation}
	\begin{aligned}
		P(x_i|x_j) & = \frac{\exp(v_i^Tv_j/\tau)}{\exp(v_j^Tv_j/\tau) + \sum_{k\ne i}\exp(v_k^Tv_j/\tau)} \\
		& = \frac{\exp(1/\tau)}{\exp(1/\tau) + \sum_{k\ne i}\exp(v_k^Tv_j/\tau)} \approx 1.\\
	\end{aligned}
	\label{eqn:softmax-diff-2}
\end{equation}

Recall our set-to-set similarity driven Softmax classification loss in Eqn.~\ref{eqn:softmax-same}, even two outliers $x_i$ and $x_j$ are identical which makes the learned distributions $\mathcal{N}_i(\mu_i, \Sigma_i)$ and $\mathcal{N}_j(\mu_j, \Sigma_j)$ indistinguishable in the early learning stage, owing to modeling of the sample uncertainty by $\Sigma_{i/j}$, the randomly sampled positive feature candidate sets $\mathcal{S}_i$ and $\mathcal{S}_j$ are not identical which will introduce more discrimination separability to $x_i$ and $x_j$. Such separability introduced by exploring uncertainty momentum can readily facilitate the loss conflict-and-collapse issue in learning and obtained more discriminative embedding.

\subsubsection{Rethinking Uncertainty Modeling from PAC-Bayesian Theory}
Furthermore, another attractive merit of our sample uncertainty modeling is multiple informative positive sample candidates can be drawn from the learned distributions for each instance. As illustrated in Fig.~\ref{F:explain}, compared with previous feature point learning-based methods~\cite{wu2018unsupervised,ye2019unsupervised,chen2020simple,he2020momentum}, the classification boundary is refined by our uncertainty modeling method so that the generalization error bound can be further reduced as demonstrated by Theorem~\ref{T:1}.

\begin{theorem}\label{T:1}
	For any unknown distribution $\mathcal{D}$, for any set $\mathcal{H}$ of classifiers, for any prior distribution $P$ on $\mathcal{H}$, for any $\delta \in (0,1]$, with probability at least $1-\delta$ over the choice of dataset $\mathcal{S} \sim \mathcal{D}$, for any posterior distribution $Q$ on $\mathcal{H}$, we have:
	\begin{equation}
		\|\mathcal{E}(\mathcal{S}, G_Q) - \mathcal{E}(\mathcal{D}, G_Q)\|^2 \le \dfrac{1}{2n}\left[D_{KL}(Q||P)+\ln\dfrac{2\sqrt{n}}{\delta}\right],
		\label{eqn:bound}
	\end{equation}
	where $\mathcal{E}(\mathcal{S}, G_Q)$ is the empirical risk over the $n$-sample learning set $\mathcal{S}$ and $\mathcal{E}(\mathcal{D}, G_Q)$ is the generalization risk over the distribution $\mathcal{D}$. $D_{KL}(Q||P)$ is the Kullback-Leibler divergence between the prior and posterior distributions.
\end{theorem}

The detailed proof is provided by \cite{begin2014pac}. From the above theorem we can see, the generalization error bound is converged to zero with a rate $O\left(\frac{\ln\sqrt{n}}{n}\right)$. Since our method can draw $k$ learning samples for each instance from the learned instance-specific distributions while previous unsupervised embedding learning methods only utilize 1 positive sample for learning, our generalization upper bound in Eqn.~\ref{eqn:bound} is much tighter which guarantees the generalizability of our proposed method.
\section{Experiment}
\label{sec:exp}
In this section, we conduct extensive experiments to demonstrate the superiority of our proposed uncertainty momentum modeling. Firstly, our method is compared with the state-of-the-art unsupervised embedding learning networks on the widely-used image classification benchmark ImageNet~\cite{russakovsky2015imagenet}; Then, following the protocol of \textit{Seen Testing Categories}~\cite{ye2019unsupervised}, two datasets (CIFAR-10 and STL-10) whose training and testing images belong to the same category are evaluated; Finally, three more benchmarks (CUB200, Car196, and Product) are tested based on the \textit{Unseen Testing Categories} protocol~\cite{ye2019unsupervised}. In addition, several ablation study experiments are conducted to thoroughly investigate our proposed method.

\subsection{Experiment on ImageNet}
\subsubsection{Implementation and Training Details}

For a fair comparison with existing unsupervised feature embedding learning methods~\cite{wu2018unsupervised,he2020momentum,chen2020improved,chen2020simple,caron2020unsupervised}, we utilize the ResNet-50 as our backbone for unsupervised embedding pretraining on ImageNet~\cite{russakovsky2015imagenet}. As we described in Sec.~\ref{sec:tech}, two branches are concatenated on the top of the output feature map from the penultimate feature extraction layer of ResNet-50 for $\mu$ and $\Sigma$ respectively. The same augmentation protocol in SimCLR~\cite{chen2020simple} is adopted by our method. For our model training on ImageNet, we use Adam optimizer~\cite{kingma2014adam} with an initial learning rate of 0.0003 for updating the ResNet-50 backbone where a weight decay of $10^{-4}$ is applied. To stabilize the training, a Cosine scheduler~\cite{loshchilov2016sgdr} is used to gradually decay the learning rate and a lower temperature $\tau=0.07$ is explored. The number of sampled positive candidates is $k=3$ for the first 100 epochs and $k=2$ for the remaining epochs in our experiments considering the trade-off between computation burden and efficiency. The weighting parameters $\lambda_{\mathcal{N}}$ and $\lambda_{\mathcal{R}}$ for $\mathcal{L}_{N}$ and $\mathcal{L}_{R}$ are chosen as 0.2 and 2 respectively to balance the scale with $\mathcal{L}_{S}$. 

Considering we just have four GeForce RTX 3090 GPUs that each one has 24GB memory, we perform two learning settings respectively: (1) we crop the image patch with a 224$\times$224 size for pretraining, batch size 128 is adopted for training 200 epochs; (2) we reduce the image patch size to a smaller one 96$\times$96, but a larger batch size 256 is used for training. By convention, when multiple GPU servers are used, the batch size is accordingly multiplied by the number of servers~\cite{hu2021adco}. As for the compared baselines (e.g, MoCo v2, SimCLR, etc), the same experimental setting is used for comparison. 

\subsubsection{Linear Classification Results}
Our proposed method, \textbf{U}ncertainty \textbf{M}omentum \textbf{M}odeling (\textbf{UMM}), is compared with other state-of-the-art unsupervised embedding learning methods including NCE~\cite{wu2018unsupervised}, LocalAgg~\cite{zhuang2019local}, MoCo~\cite{he2020momentum}, MoCo v2~\cite{chen2020improved}, PCL v2~\cite{li2020prototypical}, SimCLR~\cite{chen2020simple}, SWAV~\cite{caron2020unsupervised}). Following the same protocol in \cite{he2020momentum,chen2020simple}, a linear fully connected (FC) classifier is fine-tuned on top of the frozen 2048-D feature vector out of the pretrained ResNet-50 backbone. The linear layer is trained for 100 epochs, with a learning rate of 30 based on SGD optimizer and a batch size of 256. As reported in Table.~\ref{T:imagenet}, although we use a smaller batch size of 128, our Top@1 accuracy is comparable with the other methods which have a larger batch size of 256. Conventionally, a large batch size will result in a better linear classification result. The SWAV has a much more superior performance than all the other methods since it adopts a multi-crop strategy for performance boosting while the other methods (including ours) only use single-crop.

To further demonstrate the effectiveness of our method, we train MoCo v2, SimCLR, and our UMM using the same setting (96$\times$96 image crop and 256 batch size). The results in Table.~\ref{T:imagenet} show our UMM beats MoCo v2 and SimCLR by a significant margin (~6.4\% and ~0.6\%). Specifically, although we use a smaller image patch size for training which is more computationally efficient, our Top@1 accuracy (63.1\%) still beat SimCLR (61.9\%) who adopts a larger image patch size. The success comes from the introduced uncertainty momentum modeling.

\begin{table}[]
	\caption{\label{T:imagenet} Comparison with unsupervised embedding learning methods on ImageNet based on the \textit{FC linear classifier}. The models of pretrained 200 epochs are evaluated for all the methods.}
	\begin{center}
		\begin{tabular}{l|c|c|c|c}
			\hline
			Method 										& Patch Size & Arch ($\#$P) & Batch Size & Top@1\\
			\hline
			NCE~\cite{wu2018unsupervised}       	    & 224 & R50 (24M) & 256 & 58.5 \\
			LocalAgg~\cite{zhuang2019local}				& 224 & R50 (24M) & 128 & 58.8 \\
			MoCo~\cite{he2020momentum}       			& 224 & R50 (24M) & 256 & 60.8 \\
			SimCLR~\cite{chen2020simple}      			& 224 & R50 (24M) & 256 & 61.9 \\
			PCL v2~\cite{li2020prototypical}      		& 224 & R50 (24M) & 256 & 67.6 \\
			MoCo v2~\cite{chen2020improved}       		& 224 & R50 (24M) & 256 & 67.5 \\
			SWAV$^*$~\cite{caron2020unsupervised}       & 224 & R50 (24M) & 256 & 72.7 \\
			\textbf{UMM(Ours)}       					& 224 & R50 (24M) & 128 & 65.1 \\
			\hline
			MoCo v2~\cite{chen2020improved}       		& 96  & R50 (24M) & 256 & 62.5\\
			SimCLR~\cite{chen2020simple}      			& 96  & R50 (24M) & 256 & 56.7 \\
			\textbf{UMM(Ours)}        					& 96  & R50 (24M) & 256 & 63.1 \\
			\hline
			\multicolumn{5}{l}{*SWAV uses multi-crop for data augmentation} \\
		\end{tabular}
	\end{center}
\end{table}

\begin{table}[]
	\caption{\label{T:datasets} The statistics of testing datasets. $\#$Training/$\#$Testing/$\#$Unlabeled and $\#$Classes denote the number of training/testing/unlabeled samples and classes respectively.}
	\begin{center}
		\begin{tabular}{l|ccccc}
			\hline
			\textbf{Dataset}  & \textbf{CIFAR10} & \textbf{STL10} & \textbf{CUB200} & \textbf{Car196} & \textbf{Product}\\
			\hline
			$\#$Train	    & 50K		& 5k		& 5,864	    & 8,054	   & 59,551	\\
			$\#$Test		& 10K		& 8k		& 5,924		& 8,131    & 60,502	\\
			$\#$Unlabel	    & 0			& 100k		& 0			& 0        & 0			\\
			$\#$Classes		& 10		& 10		& 200	    & 196      & 22,634	\\
			Image Size		& 32$\times$32   	& 96$\times$96     & 256$\times$256   & 256$\times$256  & 256$\times$256		\\
			\hline
		\end{tabular}
	\end{center}
\end{table}

\begin{table}[]
	\begin{center}
		\caption{\label{T:cifar10+stl10} Comparison with unsupervised embedding learning methods on CIFAR-10 and STL-10 based on the \textit{linear classifier} and \textit{kNN classifier}.}
		\begin{tabularx}{0.45\textwidth}{XlX|XXX}
			\hline
			\multirow{2}{*}{\textbf{Linear}} & \multicolumn{2}{c|}{\textbf{CIFAR-10}} & \multicolumn{3}{c}{\textbf{STL-10}} \\
			& \textbf{Methods} & \textbf{kNN}  &  \textbf{$\#$Train} & \textbf{kNN} & \textbf{Linear}\\
			\hline
			\multirow{8}{*}{SVM}
			& DeepCluster\cite{caron2018deep}	    & 67.6 		& 5K   & 61.2 & 56.6 \\
			& NPSoftmax\cite{wu2018unsupervised} 	& 80.8      & 5K   & 66.8 & 62.3 \\
			& NCE\cite{wu2018unsupervised} 			& 80.4      & 5K   & 66.2 & 61.9 \\
			& Exemplar\cite{dosovitskiy2015discriminative} & 74.5 & 105K & N/A & 75.4 \\
			& AND~\cite{huang2019unsupervised}	    & 84.2      & 105K & 80.2 & 76.8 \\
			& ISIF\cite{ye2019unsupervised} 	    & 83.7      & 105K & 81.6 & 77.9 \\
			& PSLR~\cite{ye2020probabilistic} 		& 85.2      & 105K & 83.2 & 78.8 \\
			& \textbf{UMM(Ours)}						    & \textbf{86.3}      & \textbf{105K} & \textbf{85.5} & \textbf{83.7 }\\
			\hline
			\multirow{4}{*}{FC}
			& MoCo v2~\cite{chen2020improved}       & -         & 105K & - & 80.6 \\
			& SimCLR~\cite{chen2020simple}		    & -         & 105K & - & 76.5 \\
			& BYOL~\cite{grill2020bootstrap} 		& -         & 105K & - & 75.2 \\
			& \textbf{UMM(Ours)}				    & -         & \textbf{105K} & - & \textbf{83.6} \\
			\hline 
		\end{tabularx}
	\end{center}
\end{table}

\subsection{Experiments on CIFAR-10 and STL-10}
\subsubsection{Datasets}
Furthermore, we follow the protocol of \textit{Seen Testing Categories} in \cite{ye2019unsupervised} that two widely-used datasets (CIFAR-10~\cite{krizhevsky2009learning} and STL-10~\cite{coates2011analysis}) are tested in the experiments. The training and testing data of the two datasets share the same categories respectively and their statistic details are presented in Table.~\ref{T:datasets}.

\subsubsection{Implementation and Training Details}
We adopt a ResNet-18 network~\cite{he2016deep} as our backbone model. The training stage lasts 200 epochs and the decay rule of learning rate is that the learning rate is 0.06 for the first 60 epochs and it is decayed by 0.5, 0.1, and 0.05 at the 60, 120, and 160 epochs. As for the important hyperparameter $k$, the number of sampled positive candidates from the learned distributions, we set $k = 5$ first then $k$ is decayed to 3 and 1 at 100 and 150 epoch following the dynamic SFD rule. The temperature weight is $\tau = 0.1$, the loss weighting parameters are $\lambda_{\mathcal{N}} = 10$ and $\lambda_{\mathcal{R}} = 10$ and the embedding feature dimension is $D=128$ in experiments. The batch size is 128 for CIFAR-10 and 64 for STL-10. For all the experiments on CIFAR-10 and STL-10, a single GeForce RTX 3090 GPU is used.

\subsubsection{Compared Baselines and Evaluation Metrics}
Several state-of-the-art unsupervised deep embedding learning methods are compared including Exemplar CNN~\cite{dosovitskiy2015discriminative}, NPSoftmax~\cite{wu2018unsupervised}, NCE~\cite{wu2018unsupervised}, DeepCluster~\cite{caron2018deep}, AND~\cite{huang2019unsupervised}, ISIF~\cite{ye2019unsupervised}, MoCo v2~\cite{chen2020improved}, SimCLR~\cite{chen2020simple}, BYOL~\cite{grill2020bootstrap}, and PSLR~\cite{ye2020probabilistic}. For evaluation, we follow \cite{wu2018unsupervised,ye2019unsupervised} that a weighted kNN classifier is used for performance evaluation. Given a test sample, its top-$k$ ($k=200$) nearest neighbors based on cosine similarity are retrieved for weighted label prediction voting. Similar to the experiments on ImageNet, the linear Classification results of STL-10 are also reported. Based on the different choices of classifiers by these methods, all the methods can be categorized into two groups: using the SVM classifier and using the fully connected (FC) classifier. Our proposed method is evaluated under both two linear classifiers.

\begin{figure}[]
	\centering
	\includegraphics[width=0.45\textwidth]{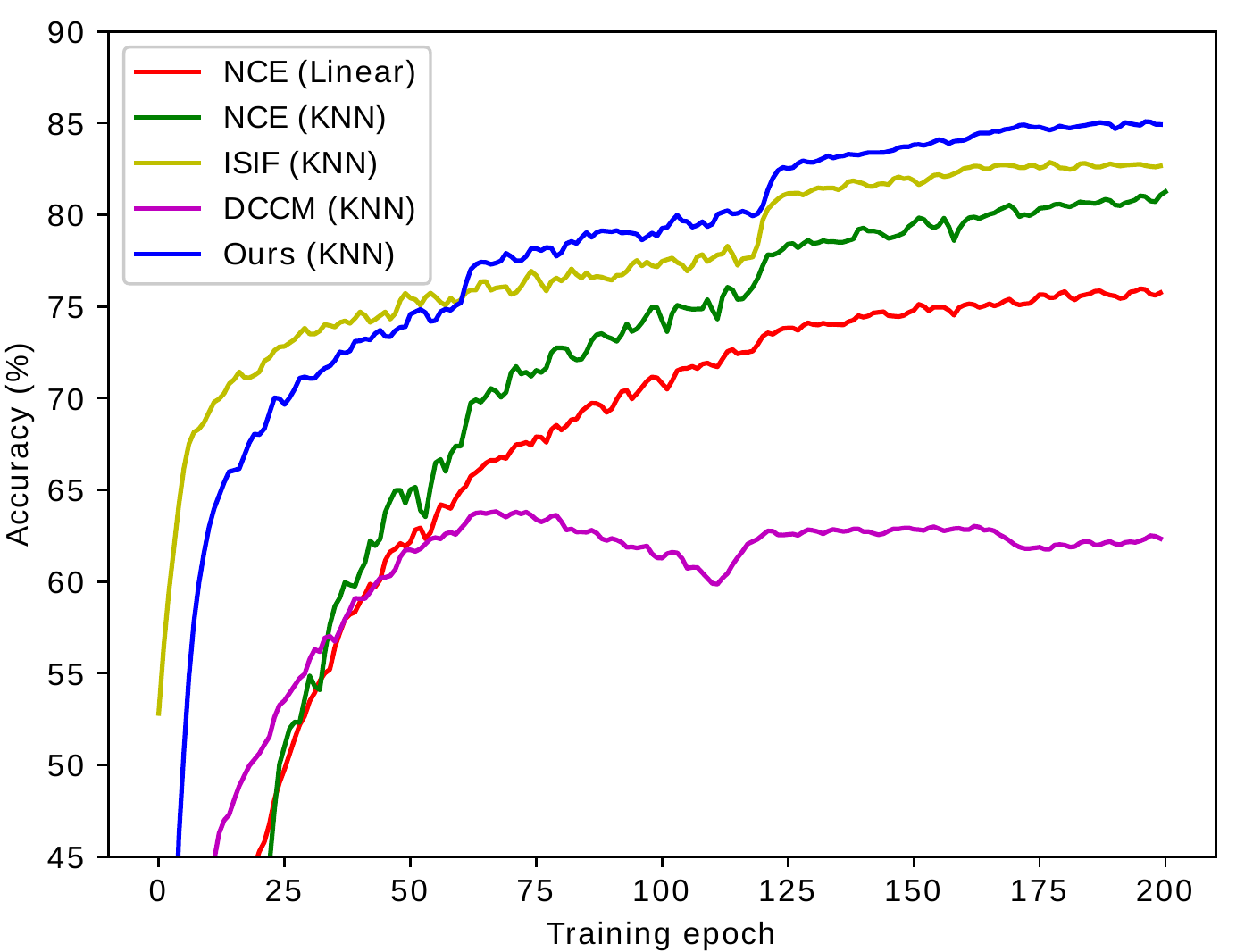}
	\caption{The kNN accuracy (\%) evaluation of different unsupervised embedding learning methods at each epoch is reported on the CIFAR-10 dataset.}
	\label{fig:epoch}
\end{figure}

\begin{table*}[!htbp]
	\caption{\label{T:cub200+product} Comparison with unsupervised embedding learning methods on CUB200, Car196 and \textit{Product} based on the image retrieval accuracy.}
	\begin{center}
		\begin{tabularx}{1\textwidth}{lXXXX|XXXX|XXXX}
			\hline
			& \multicolumn{4}{c}{\textbf{CUB200}} & \multicolumn{4}{c}{\textbf{Car196}} & \multicolumn{4}{c}{\textbf{Product}} \\
			\hline
			\textbf{Methods} & R@1 & R@2 & R@4 & NMI & R@1 & R@2 & R@4 & NMI & R@1 & R@10 & R@100 & NMI \\
			\hline
			Initial(FC) 								    & N/A  & N/A  & N/A  & N/A  & 35.1 & 47.4 & 60.0 & 38.3 & 40.8 & 56.7 & 72.1 & 84.0 \\
			Cyclic\cite{caron2018deep} 						& 40.8 & 52.8 & 65.1 & 52.6 & N/A  & N/A  & N/A  & N/A  & N/A  & N/A  & N/A  & N/A \\
			Exemplar\cite{dosovitskiy2015discriminative} 	& 38.2 & 50.3 & 62.8 & 45.0 & 36.5 & 48.1 & 59.2 & 35.4 & 45.0 & 60.3 & 75.2 & 85.0 \\
			NCE\cite{wu2018unsupervised} 					& 39.2 & 51.4 & 63.7 & 45.1 & 37.5 & 48.7 & 59.8 & 35.6 & 46.6 & 62.3 & 76.8 & 85.8 \\
			DCluster\cite{caron2018deep}					& 42.9 & 54.1 & 65.6 & 53.0 & 32.6 & 43.8 & 57.0 & 38.5 & 34.6 & 52.6 & 66.8 & 82.8 \\
			MOM\cite{iscen2018mining} 						& 45.3 & 57.8 & 68.6 & 55.0 & 35.5 & 48.2 & 60.6 & 38.6 & 43.3 & 57.2 & 73.2 & 84.4 \\
			ISIF\cite{ye2019unsupervised} 			        & 46.2 & 59.0 & 70.1 & 55.4 & 40.1 & 50.8 & 62.6 & 34.3 & 48.9 & 64.0 & 78.0 & 86.0 \\
			\hline
			\textbf{UMM(Ours)}							        & \textbf{47.3} & \textbf{59.5} & \textbf{70.6} & \textbf{56.0} & \textbf{40.5} & \textbf{52.4} & \textbf{63.3} & \textbf{37.3} & \textbf{49.4} & \textbf{65.1} & \textbf{78.7} & \textbf{87.6}\\
			\hline 
		\end{tabularx}
	\end{center}
\end{table*}

\subsubsection{Results}
Table.~\ref{T:cifar10+stl10} demonstrates the kNN classification performance based on the learned feature embedding from different methods on CIFAR-10. Our proposed method outperforms the others by a large margin, from 85.2\% to 86.3\% on CIFAR-10. Compared with the state-of-the-art baseline PSLR, the significant improvement of our method benefits from modeling the sample distribution uncertainty and the learning based on set-to-set similarity relationships. As for another large-scale STL-10 dataset which contains 105K training images in which 100K are unlabeled images but share a similar distribution with the 5K labeled images for unsupervised learning, our proposed method still improves the state-of-the-art significantly (from 83.2\% to 85.5\% at kNN accuracy) which is able to demonstrate the scalability and superiority of our method. Moreover, the \textit{Linear} classifier which requires additional classifier learning with the labeled training data is also evaluated for the learned embedding features. Our proposed method achieves a state-of-the-art performance that beats PSLR and MoCo v2 significantly on linear classification accuracy (from 78.8\% to 83.7\% on SVM classifier, and from 80.6\% to 83.6\% on FC classifier).

To provide a complete comparison, the learning curves of the compared state-of-the-art unsupervised embedding learning methods plus ours at different learning epochs on CIFAR-10 are plotted in Fig.~\ref{fig:epoch}. As we can see, for the first 50 epochs, our performance is slightly lower than ISIF~\cite{ye2019unsupervised} since the uncertainty of samples in the initial distribution learning phase is pretty large which requires more learning epochs to stabilize the distribution learning. While our method outperforms all the other rivals by a large margin after 50 epochs which demonstrates the effectiveness of modeling the uncertainty of samples via distribution learning.

\begin{figure*}[]
	\begin{center}
		\includegraphics[width=0.95\textwidth]{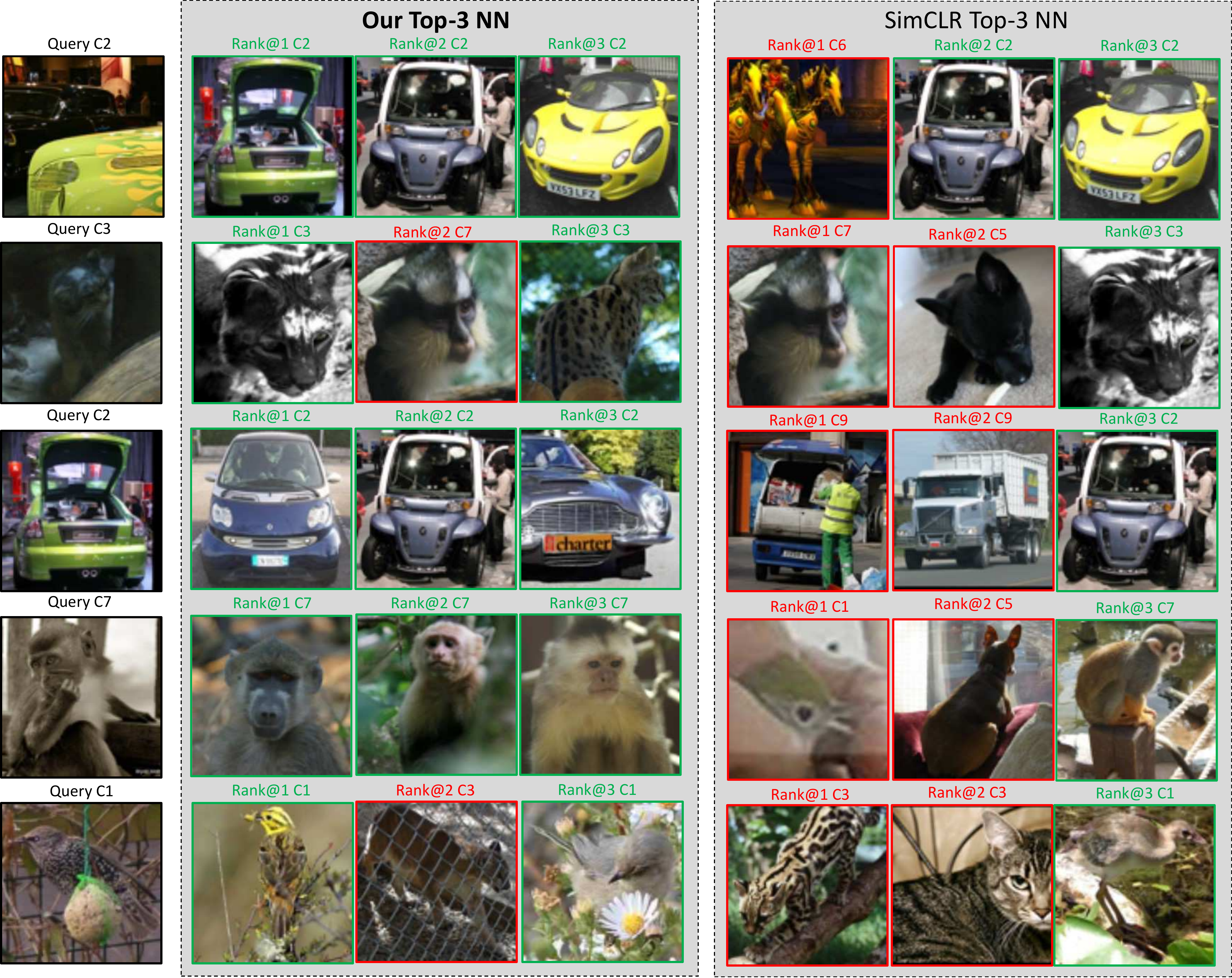}
	\end{center}
	\caption{\label{F:visualization_outlier} Given a learning minibatch of STL-10, the Top-3 nearest neighbors of a query instance are retrieved from the same minibatch based on the extracted feature representations from the SimCLR network and our method. (e.g., Rank@1 C6 means this instance is the top-1 NN of the query instance and its class label is 6.) Each row represents one retrieval result of a query case. As we can see, the SimCLR suffers from the critical outliers which are visually similar but belong to totally different classes. While our method can successfully push these outliers farther away so that the discrimination of the query instances is better enhanced.}
\end{figure*}

\begin{figure}[]
	\begin{center}
		\subfloat[Loss Components]{\includegraphics[width=0.4\textwidth]{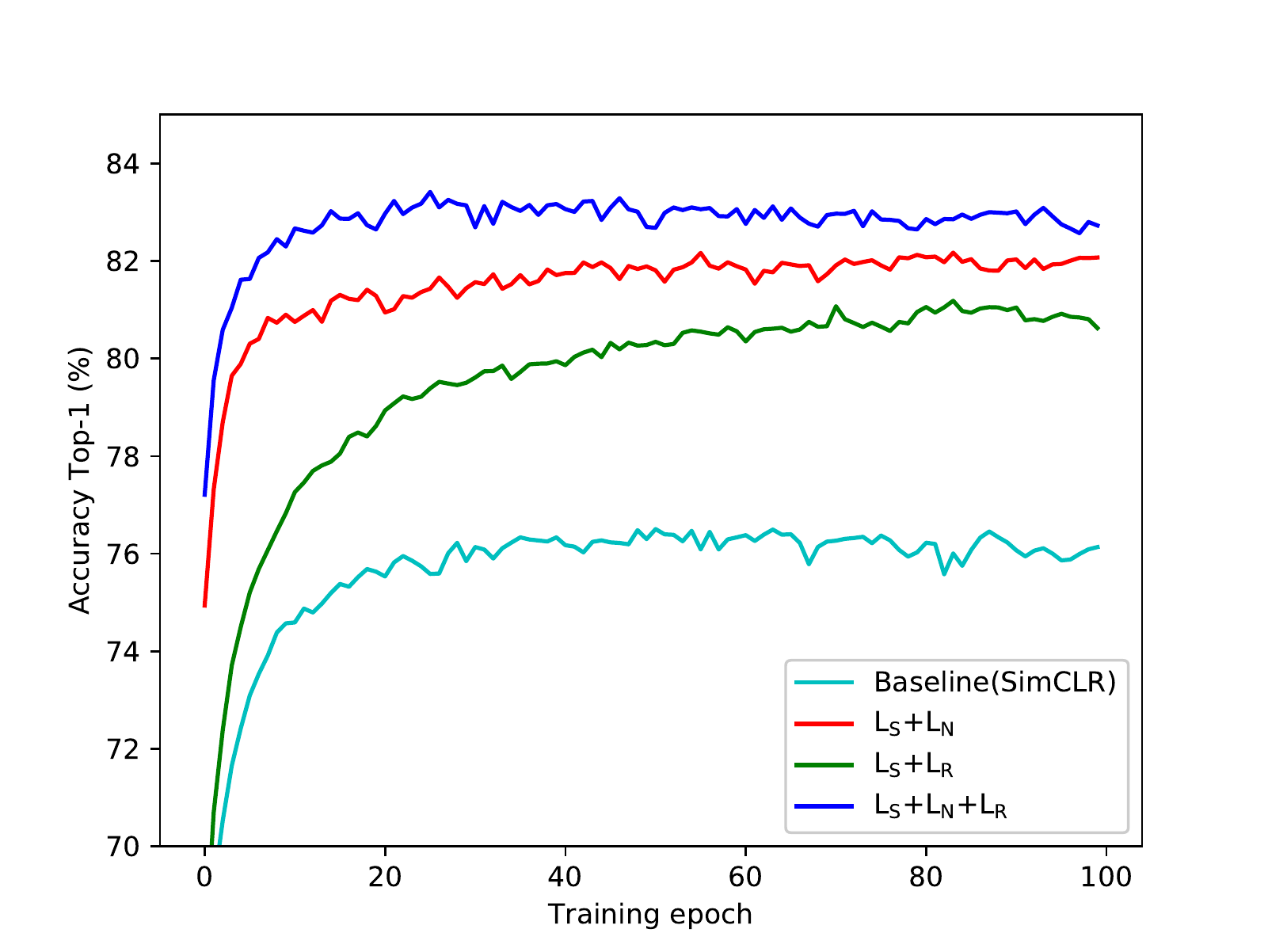}}
		\hfil
		\subfloat[$\lambda_{\mathcal{N}}$=$\lambda_{\mathcal{R}}$=$\lambda$]{\includegraphics[width=0.4\textwidth]{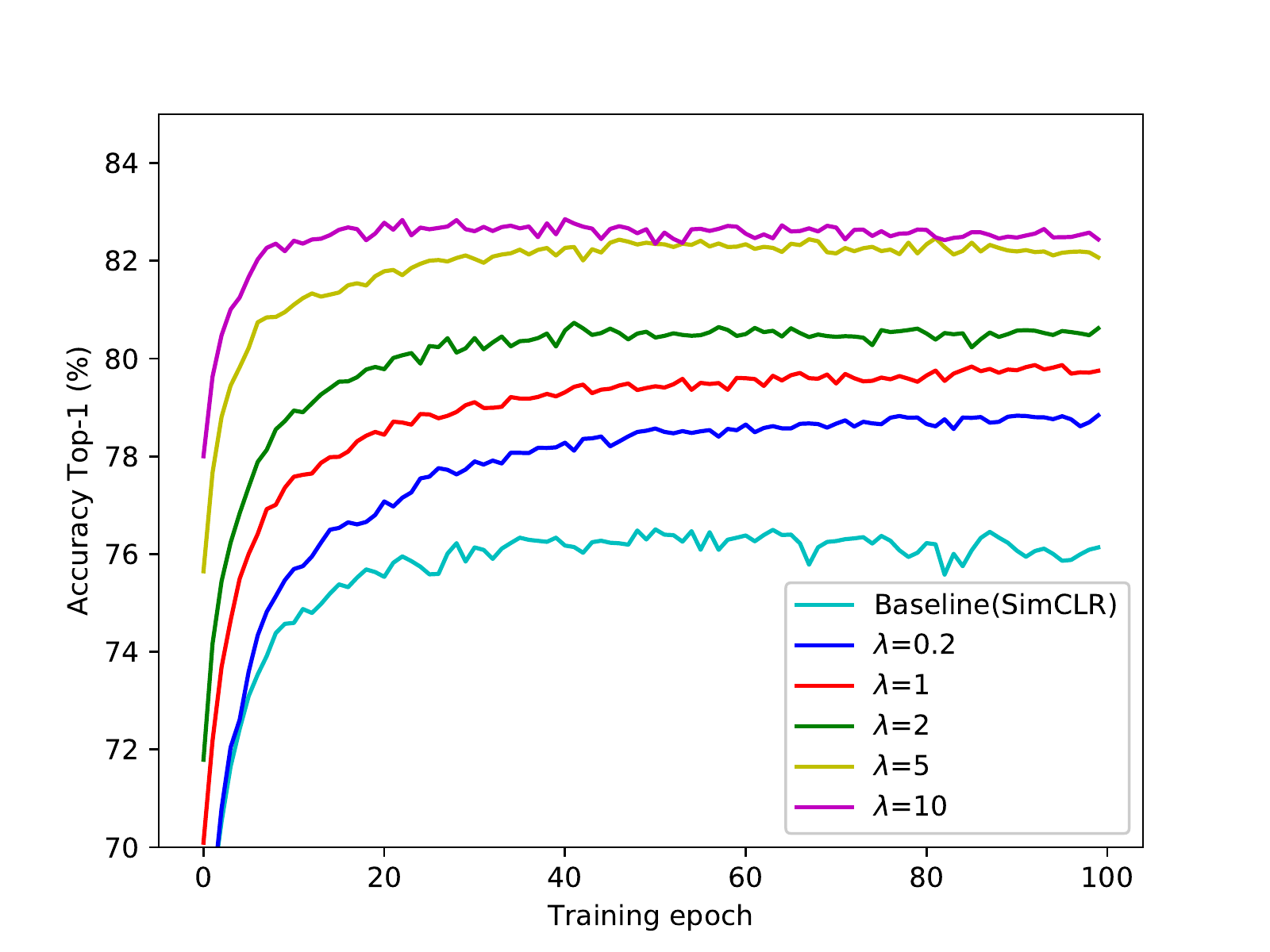}}
		\hfil
		\subfloat[Sampling Number $k$]{\includegraphics[width=0.4\textwidth]{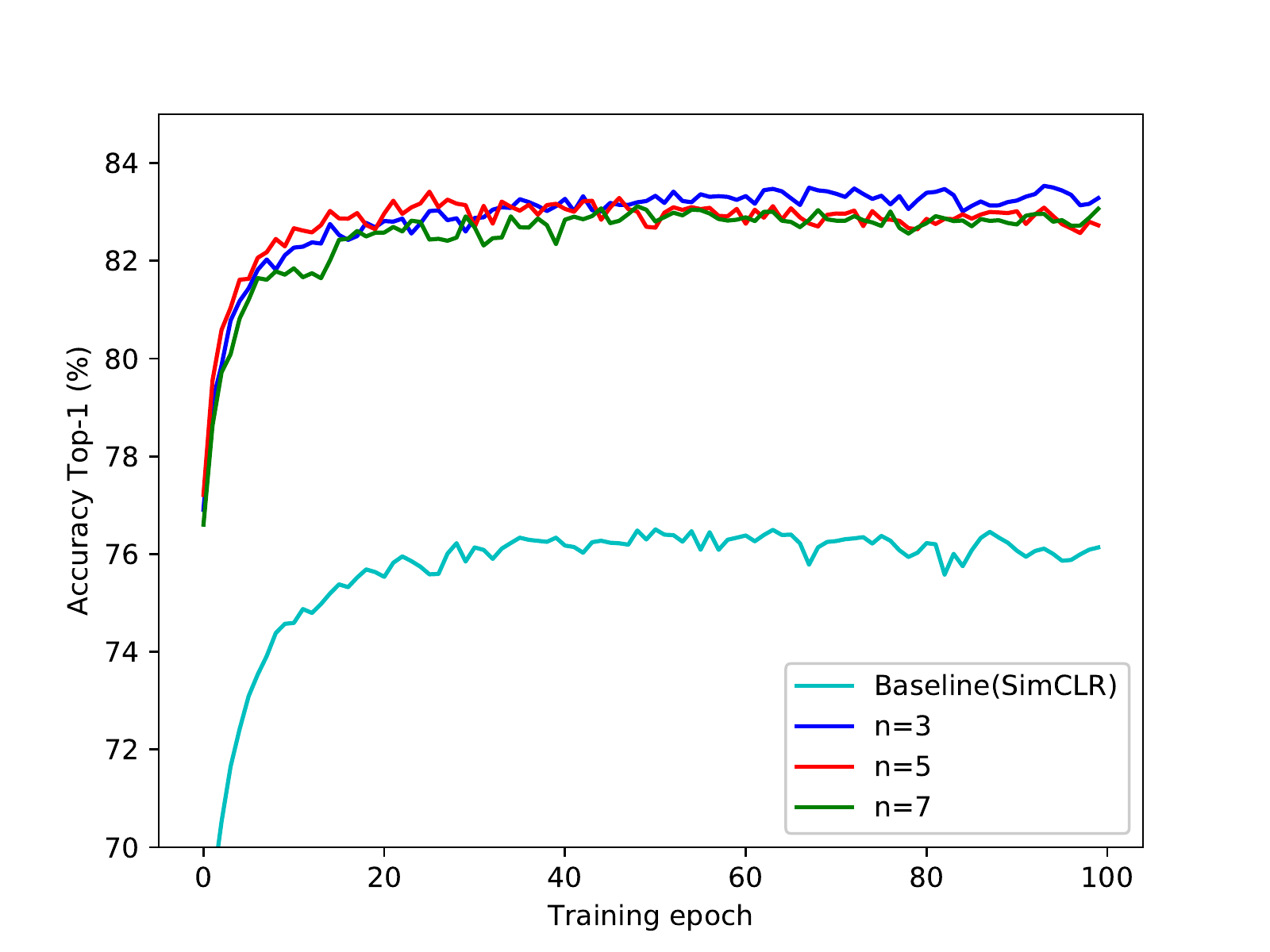}}
	\end{center}
	\caption{\label{fig:ablation-stl10} The visualization of linear classification (FC classifier) accuracy (Top@1) under different experimental parameters on STL-10.}
\end{figure}

\begin{figure}[]
	\begin{center}
		\subfloat[ISIF~\cite{ye2019unsupervised} (D=128)]{\includegraphics[width=0.25\textwidth]{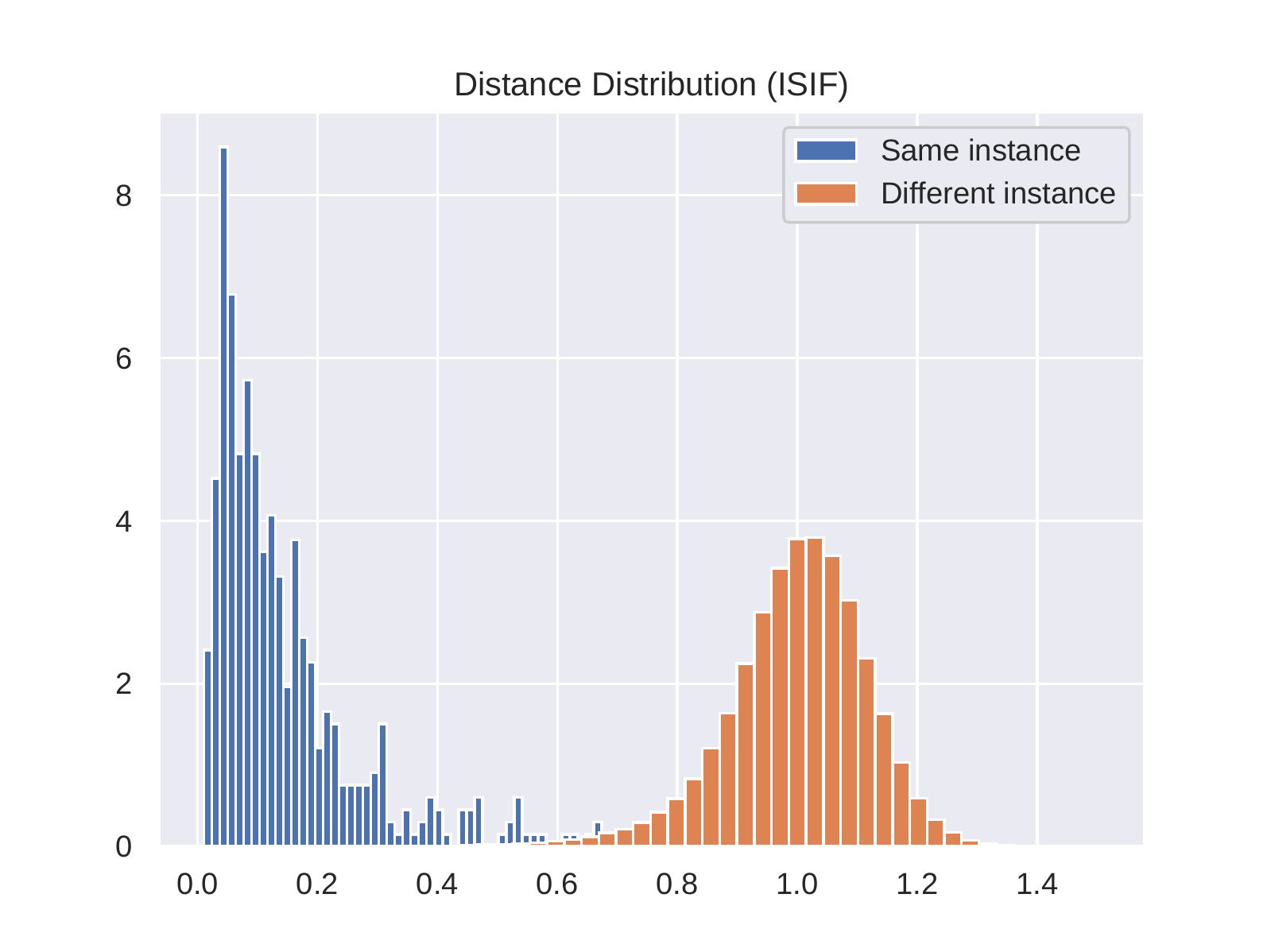}}
		\subfloat[UMM (D=128)]{\includegraphics[width=0.25\textwidth]{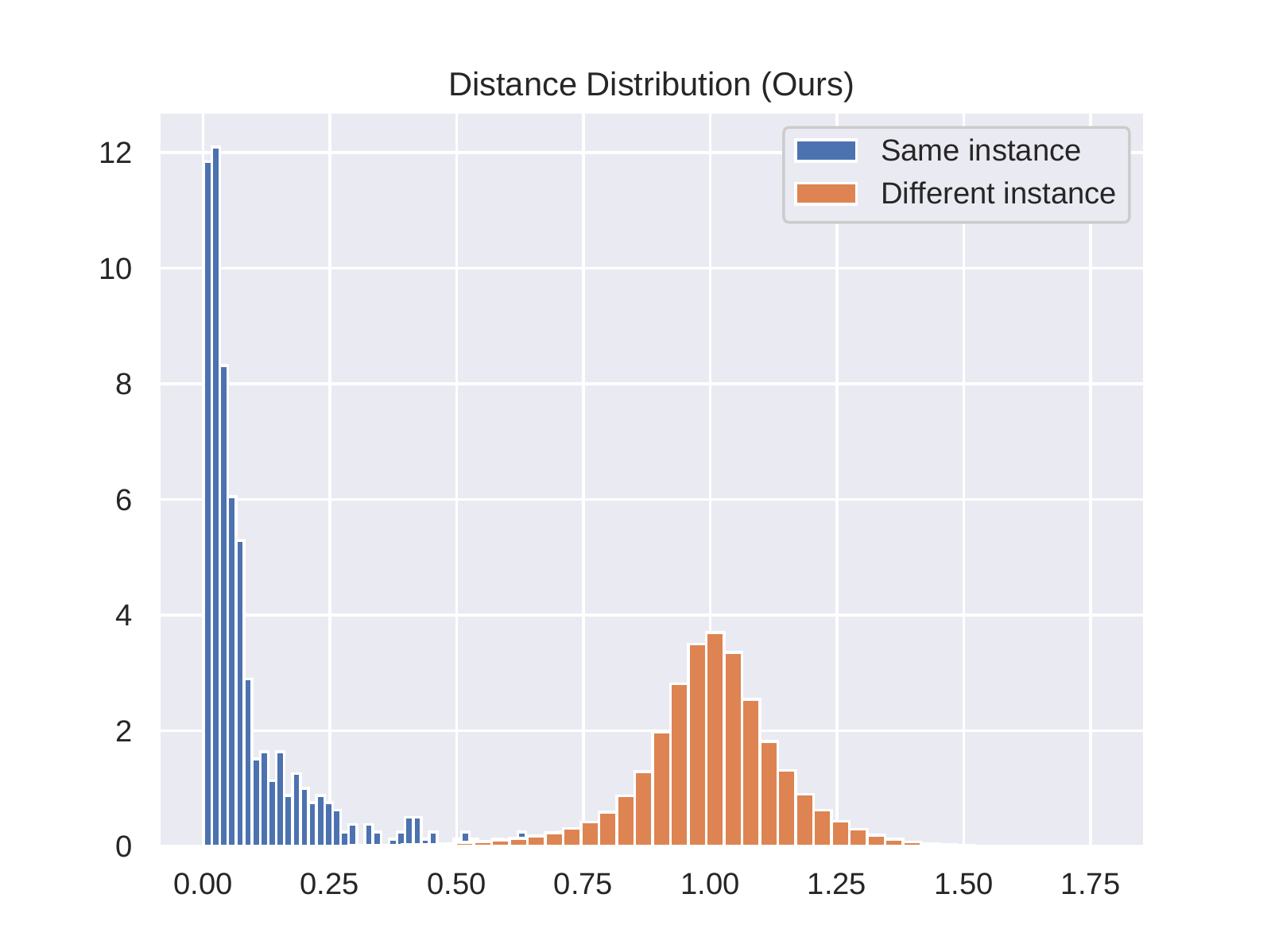}}
		\hfill
		\subfloat[ISIF~\cite{ye2019unsupervised} (D=512)]{\includegraphics[width=0.25\textwidth]{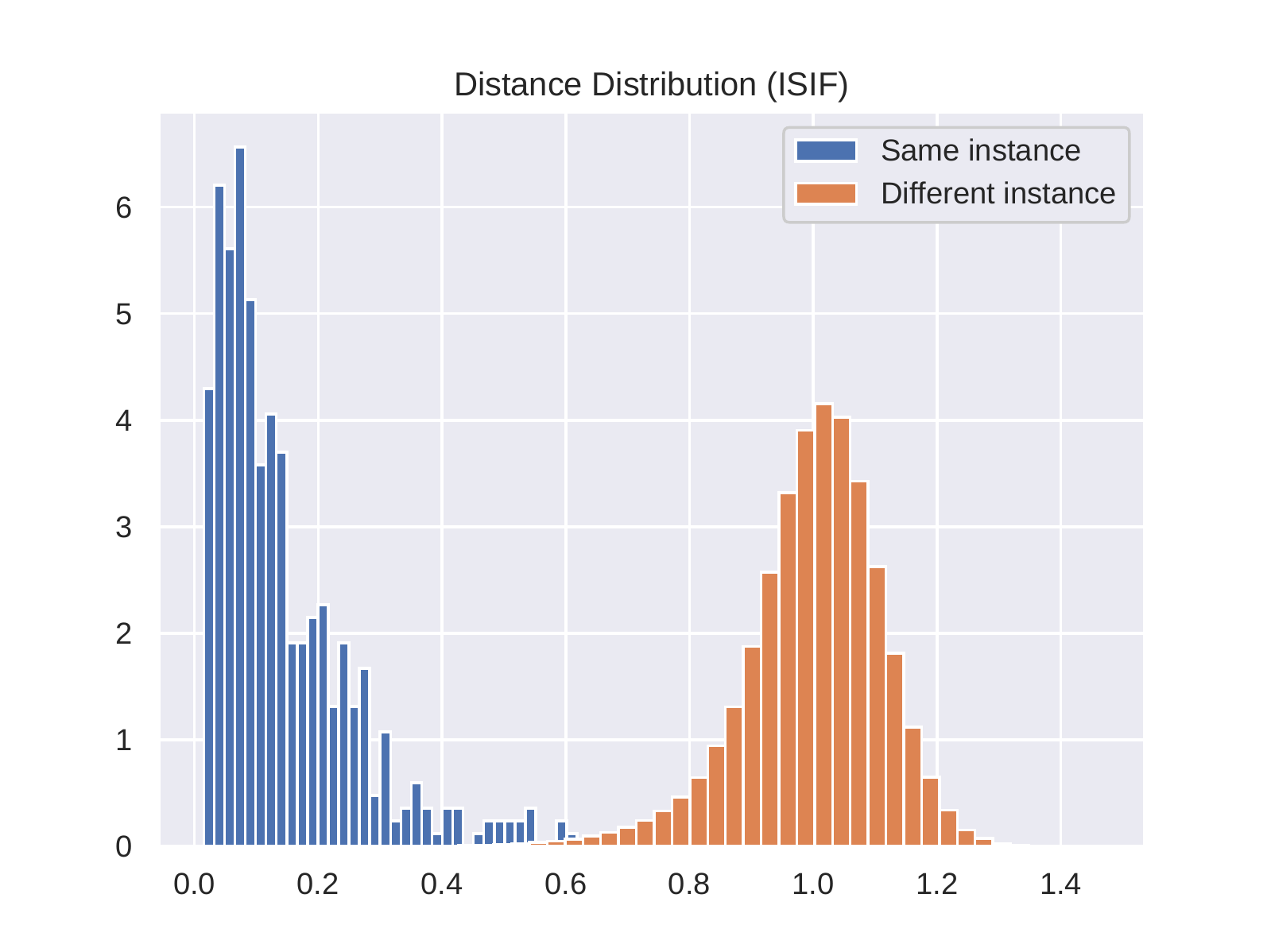}}
		\subfloat[UMM (D=512)]{\includegraphics[width=0.25\textwidth]{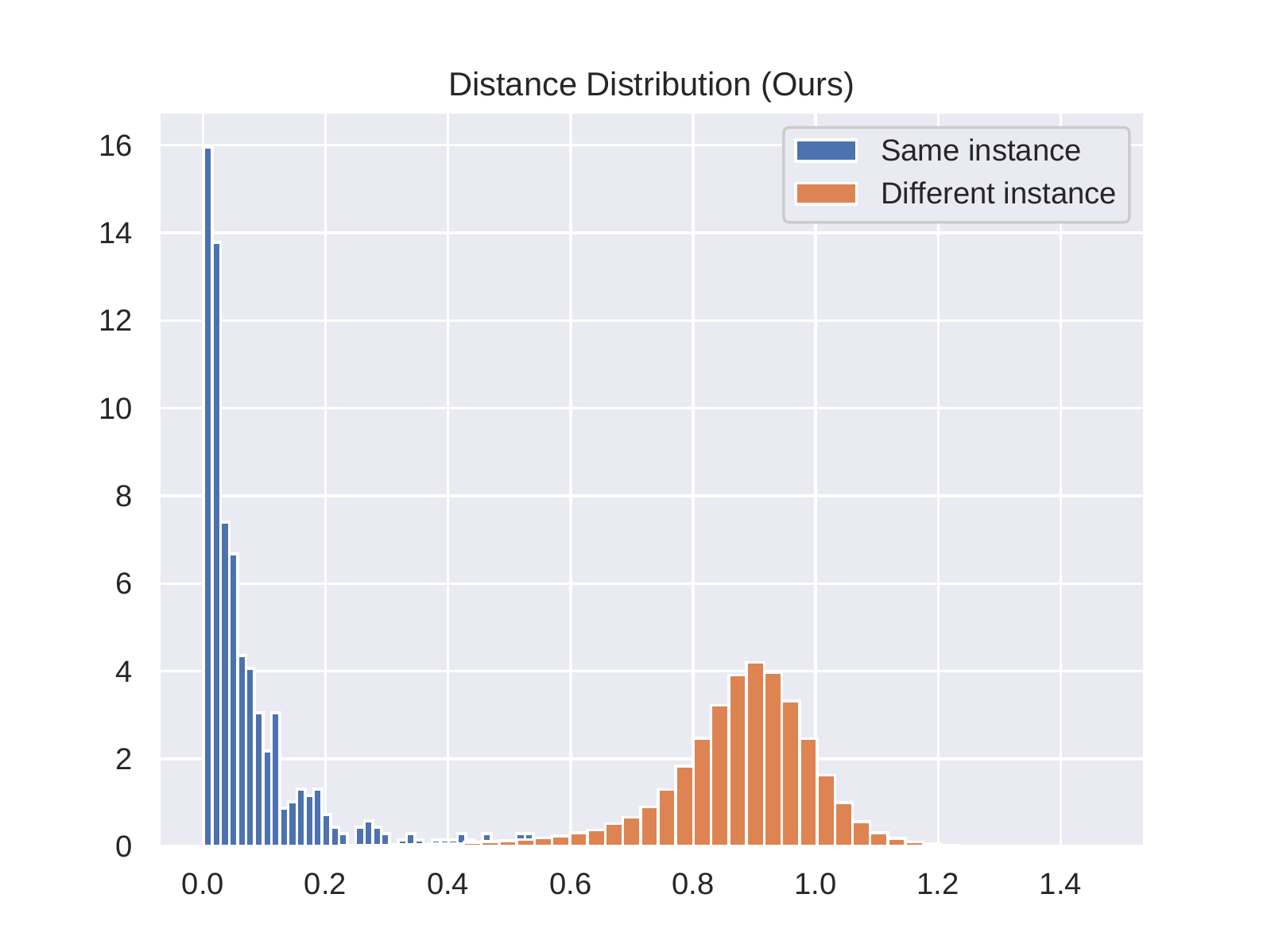}}
	\end{center}
	\caption{\label{F:visualization_distribution} The cosine similarity distributions of ISIF~\cite{ye2019unsupervised} and our UMM on CIFAR-10 under different feature dimensions ($D=128$ and $D=512$).}
\end{figure}

\begin{figure*}[]
	\begin{center}
		\subfloat[Epoch-0]{\includegraphics[width=0.25\textwidth]{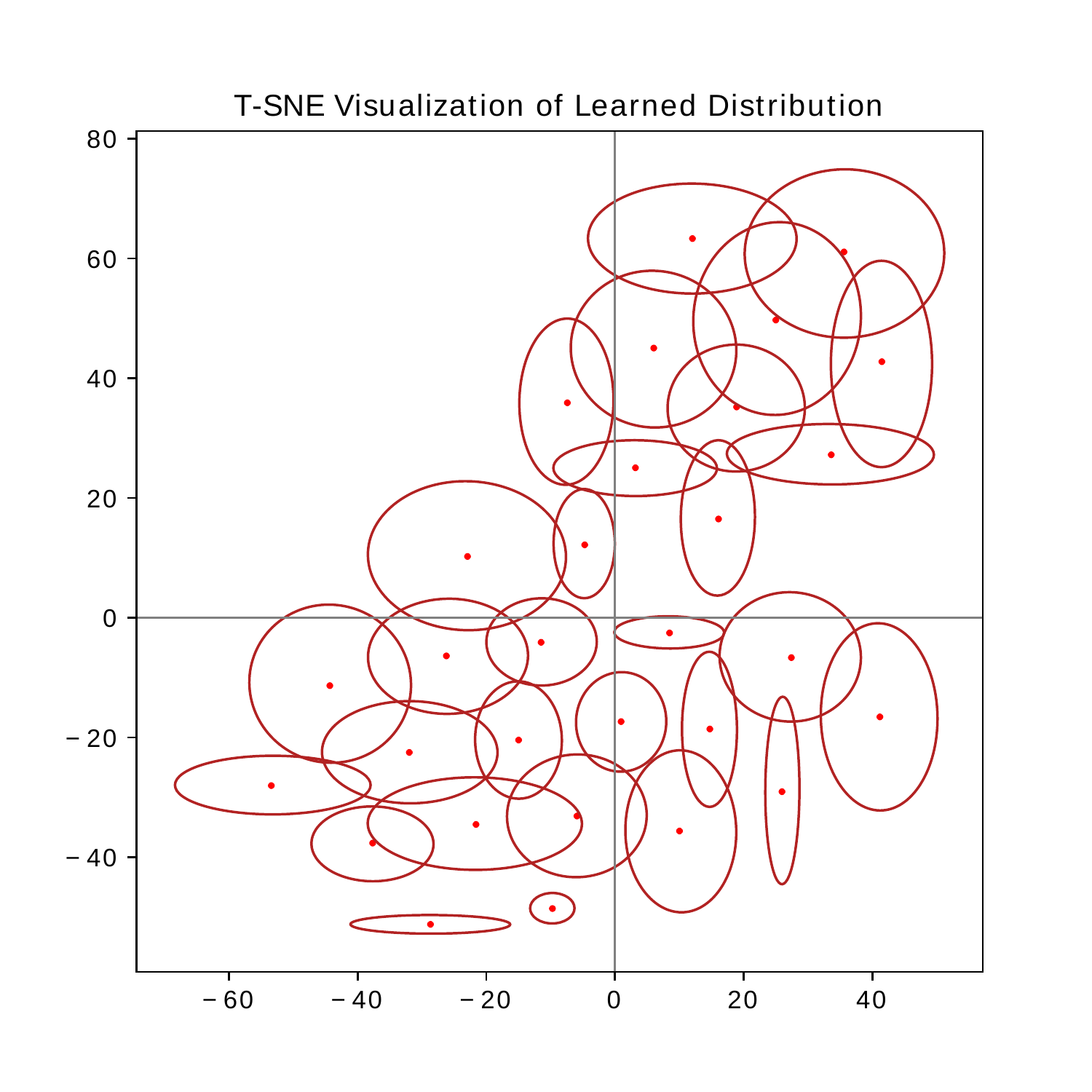}}
		\subfloat[Epoch-10]{\includegraphics[width=0.25\textwidth]{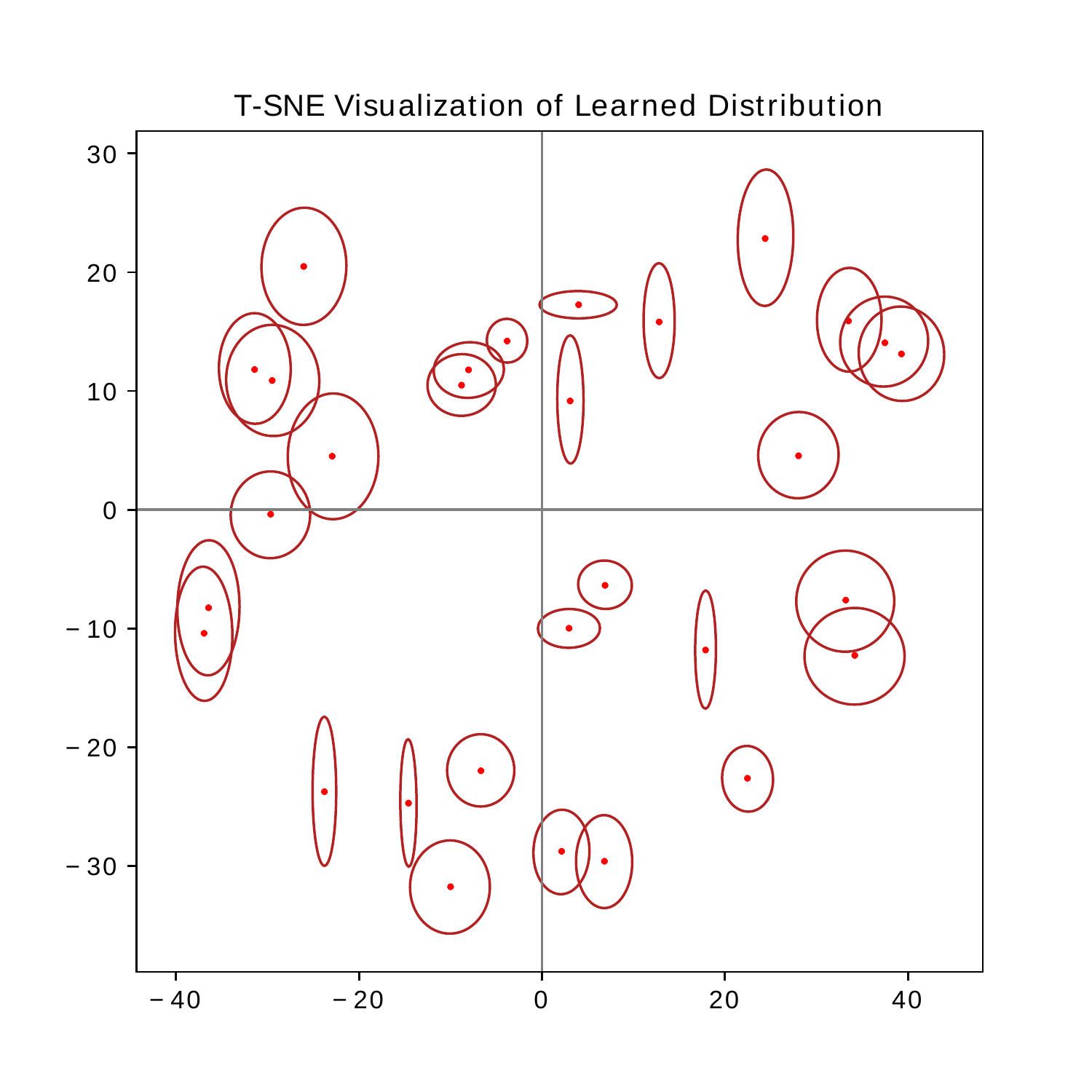}}
		\subfloat[Epoch-100]{\includegraphics[width=0.25\textwidth]{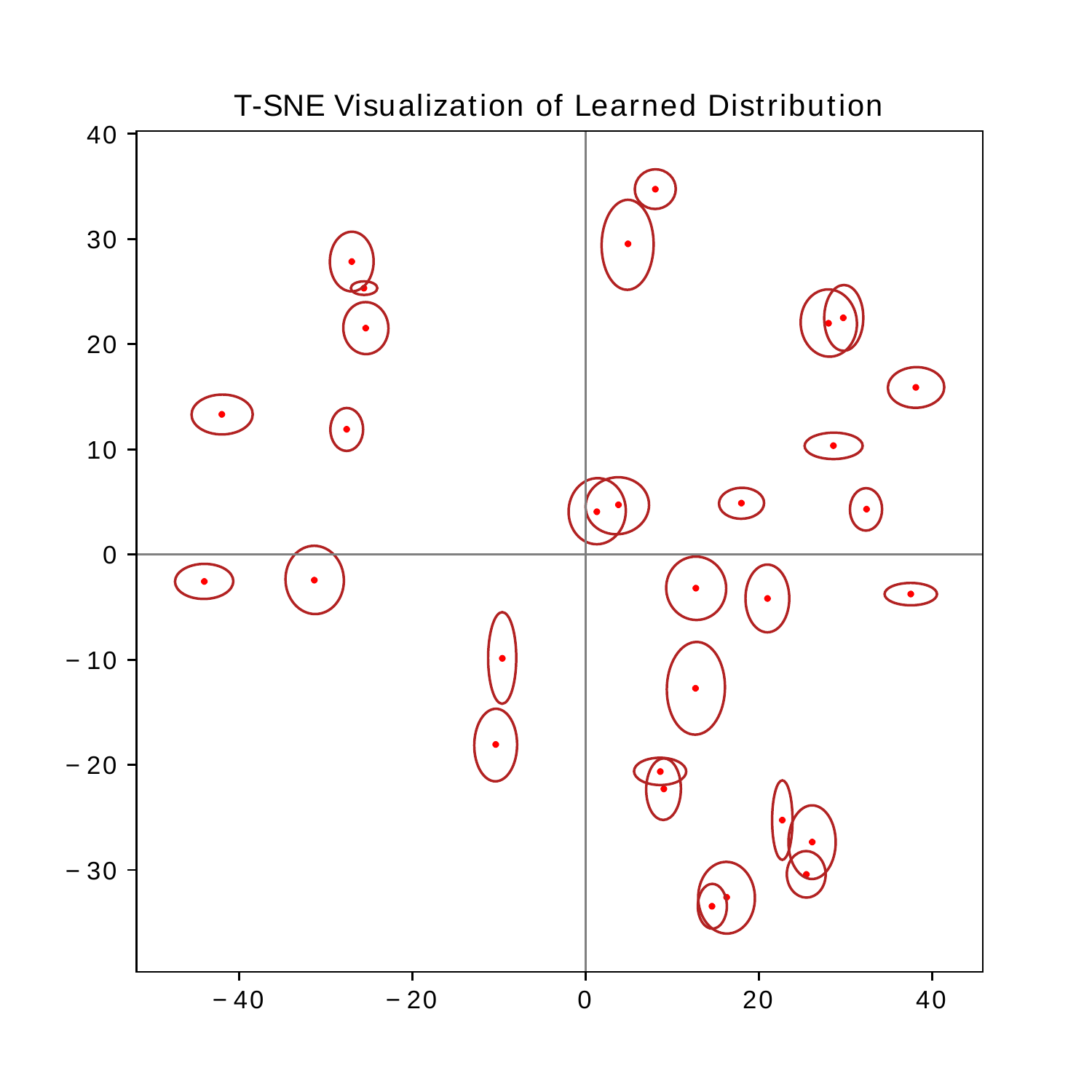}}
		\subfloat[Epoch-200]{\includegraphics[width=0.25\textwidth]{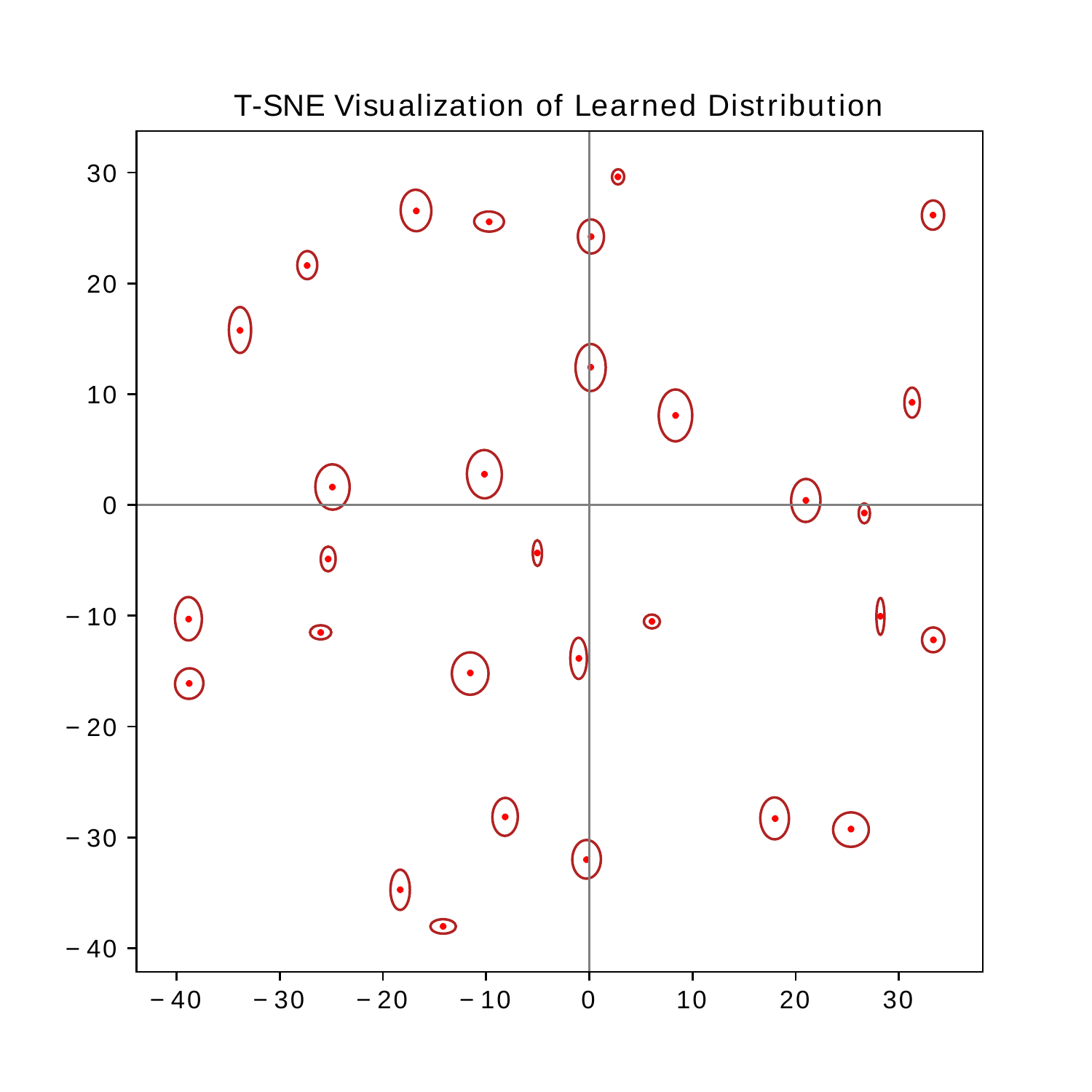}}
	\end{center}
	\caption{\label{F:visualization_learning} The visualization of the learned instance-specific distributions on CIFAR-10 at different training stages (Epoch-0, Epoch-10, Epoch-100 and Epoch-200). Each red dot is the center localization ($\mu$) and the circle range represents the uncertainty ($\Sigma$).}
\end{figure*}

\subsection{Experiments on CUB200, Car198 and Product}
\subsubsection{Datasets}
Three more datasets, CUB200-2011 (CUB200)~\cite{wah2011caltech} Car196~\cite{krause20133d}, and Stanford Online Product (\textit{Product})~\cite{oh2016deep}, are evaluated in our experiments whose details are listed in Table.~\ref{T:datasets}. For these three datasets, the training and testing samples are drawn from completely different classes so that they are adopted for the evaluation under the  setting of \textit{Unseen Testing Categories}. More specifically, the training sets of CUB200, Car196, and \textit{Product} contain samples from 100, 96, and 11,318 classes respectively and the samples in their testing sets are collected from the other 100, 96, and 11,316 different classes respectively.

\subsubsection{Implementation and Training Details}
We adopt the pre-trained Inception-V1~\cite{szegedy2015going} model on ImageNet as our backbone network~\cite{sohn2016improved,oh2016deep}. Two FC layers with $l_2$ normalization are added after the pool5 layer as the distribution learning layers. The learning rate is set to 0.001 without decay and $k$ is set to 50 firstly then decayed to 30 and 10 after 30 and 60 epochs. The temperature weight $\tau = 0.1$, batch size 64 and the embedding feature dimension $D=128$ are adopted. The loss weighting parameters are $\lambda_{\mathcal{N}} = 0.1$ and $\lambda_{\mathcal{R}} = 0.1$. For all the experiments on these three datasets, a single GeForce RTX 3090 GPU is used.

\subsubsection{Compared Baselines and Evaluation Metrics}
Similarly, the compared methods in CIFAR-10 and STL-10 datasets are also evaluated here. As for the CUB200, Car196 and \textit{Product} datasets, we follow the same protocol in \cite{ye2019unsupervised} that the retrieval performance (Rank@k) and clustering quality (NMI) of the testing set are evaluated.

\subsubsection{Results}
The effectiveness and superiority of modeling instance-level feature uncertainty are not only demonstrated under the \textit{Seen Testing Category} scenario (CIFAR-10 and STL-10), but also verified by the following experiments under \textit{Unseen Testing Category} setting. On all three benchmarks, our proposed method significantly outperforms the second-best player on all Rank@k criteria which demonstrates the discriminative ability of our learned feature embedding. For the state-of-the-art methods NCE~\cite{wu2018unsupervised} and ISIF~\cite{ye2019unsupervised}, although the learned features perform well on retrieval evaluation, their clustering quality is pretty poor compared with other clustering-based methods, DeepCluster~\cite{caron2018deep} and MOM\cite{iscen2018mining}. While our proposed method not only beats all of them on the retrieval performance but also greatly outperforms NCE~\cite{wu2018unsupervised} and ISIF~\cite{ye2019unsupervised} under NMI evaluation especially on the challenging Car196 dataset, obtaining a competitive result against the clustering-based models. Such experiment results demonstrate that our learned feature embedding is much more consistent with the inherent sample distribution owing to the modeling of instance feature uncertainty modeling.

\subsection{Ablation Study}
For the sake of computation convenience, extensive ablation studies are conducted on four widely-used datasets (CIFAR-10, STL-10, CUB200, and Car196) to thoroughly investigate our proposed model. As demonstrated by Sec.~\ref{sec:tech}, several parameters influence our model performance including different combinations of loss components, the number of sampled positive candidates $k$, the embedding dimension $D$, and so on.

\begin{figure*}[]
	\begin{center}
		\includegraphics[width=1\textwidth, height=0.25\linewidth]{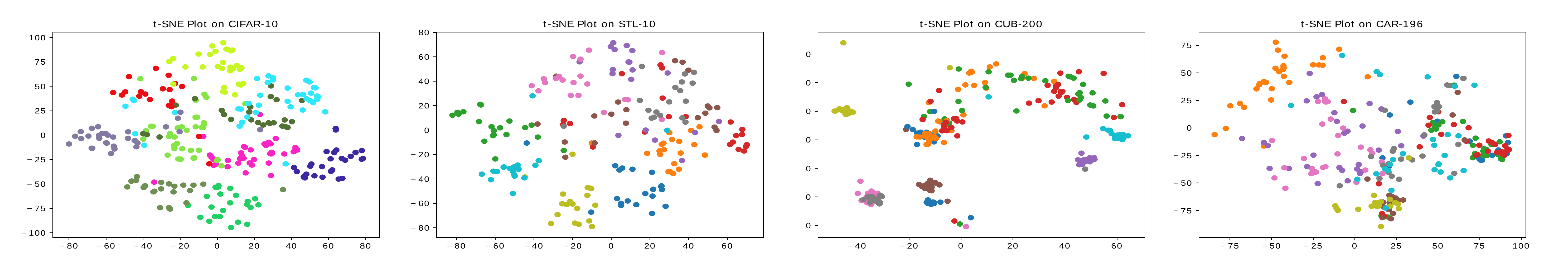}
	\end{center}
	\caption{\label{F:visualization_tsne} The T-SNE visualization of our learned embedding space of testing samples on CIFAR-10, STL-10, CUB200, and Car196 respectively. (200 samples are randomly selected)}
\end{figure*}

\textbf{The improvement of sample outliers via our uncertainty modeling}: To verify that the improvement of our method indeed comes from the well-tackling of sample outliers, we conduct the following experiments: For the unlabeled learning images in a minibatch of STL-10, the corresponding feature descriptors are extracted based on SimCLR network and our method. As presented by Fig.~\ref{F:visualization_outlier}, given such a learning minibatch of STL-10, the Top-3 nearest neighbors of a query instance are retrieved from the same minibatch based on the extracted feature representations from the SimCLR network and our method. (For example, \textit{Rank@1 C6} means this instance is the top-1 NN for the query and its class label is 6.) As we can see, the SimCLR suffers from the critical outliers which are visually similar but belong to totally different classes. While our UMM method can successfully push these outliers farther away as well as improve the rank of same-category instances so that the discrimination of the query instances is better enhanced. This result verifies that our performance improvement is mainly achieved by the better handling of these unlabeled outliers during learning.

\begin{table}[]
	\caption{The ablation study of our method on CIFAR-10 dataset.}
	\centering
	\begin{tabular}{l|lc}
		\hline
		\textbf{Settings} & \textbf{Methods} & \textbf{kNN} \\
		\hline
		\multirow{5}{*}{Loss Component}
		& Baseline\cite{ye2019unsupervised} & 83.7 \\
		& $\mathcal{L}_S$ & 84.9 \\
		& $\mathcal{L}_S$ + $\mathcal{L}_{\mathcal{N}}$ & 85.8\\
		& $\mathcal{L}_S$ + $\mathcal{L}_{\mathcal{R}}$ & 85.2 \\
		& $\mathcal{L}_S$ + $\mathcal{L}_{\mathcal{N}}$ + $\mathcal{L}_{\mathcal{R}}$ & 86.3 \\
		\hline
		\multirow{4}{*}{Embedding Dim}
		& $D=128$     & 86.3 \\
		& $D=256$     & 86.7 \\
		& $D=512$     & 87.3 \\
		& $D=1024$    & 86.9 \\
		\hline
		\multirow{4}{*}{Sampling Number}
		& $k=1$       & 85.3 \\
		& $k=3$       & 86.1 \\
		& $k=5$       & 86.3 \\
		& $k=10$      & 85.2 \\
		\hline
	\end{tabular}
	\label{T:ablation}
\end{table}

\textbf{The influence of loss components}: As presented by Eqn.~\ref{eqn:final_loss}, our overall learning loss contains three different components which play different roles in the learning. As reported in Table.~\ref{T:ablation}, by only keeping the set-to-set distance-based Softmax classification loss $\mathcal{L}_{S}$, our method is still able to beat the baseline model ISIF~\cite{ye2019unsupervised} by 1.2\% on kNN accuracy. In addition, the effectiveness of replacing the point-to-point similarity measurement to a set-to-set one is further demonstrated by combining the distribution consistency loss $\mathcal{L}_{\mathcal{N}}$ with $\mathcal{L}_{S}$, a significant improvement (over 0.9\%) can be further obtained. The global ranking loss $\mathcal{L}_{\mathcal{R}}$ can also introduce a non-trivial improvement to our model and make the learning much more stable. Finally, learning using our overall loss performs the best. A similar result can also be observed by the linear classification results on STL-10. As shown by Fig.~\ref{fig:ablation-stl10}(a), the linear classifier that is trained based on the learned feature embedding from the overall loss achieves the best classification performance. Besides, learning from any one of the three loss components can outperform the baseline model (SimCLR) by a large margin.

\textbf{The influence of weighting parameters $\lambda_{\mathcal{N}}$ and $\lambda_{\mathcal{R}}$}: The linear classification results shown in Fig.~\ref{fig:ablation-stl10}(b) demonstrate the influence of the weighting parameters $\lambda_{\mathcal{N}}$ and $\lambda_{\mathcal{R}}$ for our proposed losses. As we can see, increasing $\lambda_{\mathcal{N}}$ and $\lambda_{\mathcal{R}}$ results in a improvement of final performance since the learned instance distribution is more accurate ($\mathcal{L}_{\mathcal{N}}$) which can provide better positive candidates for the global discrimination enhancement ($\mathcal{L}_{\mathcal{R}}$). Finally, $\lambda_{\mathcal{N}}=10$ and $\lambda_{\mathcal{R}}=10$ achieves the best performance on STL-10.

\textbf{The influence of embedding dimension $D$}: Existing unsupervised embedding learning methods propose to learning a relatively low-dimensional ($D=128$) feature embedding, and the experiment results in \cite{ye2019unsupervised} show even the feature dimension is increased to $D=512$, the state-of-the-art methods can not benefit from it. However, our UMM reports a further and significant improvement (from 86.3\% to 87.3\%) on CIFAR-10 by increasing feature dimension from 128 to 512 which demonstrates the dimensional scalability of our sample uncertainty modeling.

\textbf{The influence of sampled positive candidate number $k$}: The number of sampled positive candidates from the learned distribution is a crucial hyperparameter in our method since $k$ controls the quality of learning performance. If $k$ is too small in the early learning phase, the sampled positive candidate set $\mathcal{Z}_i$ is not enough to depict the uncertainty, while too large $k$ will occupy much memory storage and severely slow down the learning process. As demonstrated by the ablation study results in CIFAR-10 (Table.~\ref{T:ablation}) and STL-10 (Fig.~\ref{fig:ablation-stl10}(c)), $k=1$ means the set-to-set distance in Eqn.~\ref{eqn:set-to-set distance} will degenerate to a point-to-point similarity measurement then the randomness of data sampling will largely influence the learning performance. By gradually increasing $k$ to 5, much better learning performance will be obtained which presents that only sampling a small number of positive candidates is already helpful for instance discrimination enhancement. When we keep increasing $k$, only minor improvement or even performance degradation happens because many redundant candidates are extracted but can not provide extra useful information for learning.

\textbf{The convergence of uncertainty modeling}: The necessity of our proposed dynamic sampling-from-distribution strategy relies on the assumption that the uncertainty of samples will become smaller as the learning epoch increases. To verify this, we visualize the learned distributions of given unlabeled samples on CIFAR-10 at different training stages. As illustrated by Fig.~\ref{F:visualization_learning}, the learned distributions at the initial stage (Epoch-0) show pretty large uncertainty. When learning goes to epochs 10, 100, and 200, the uncertainty of samples is gradually and consistently reduced which makes the discriminative capability of our model enhanced.

\textbf{The visualization of cosine similarity distributions on CIFAR-10}: The distributions of cosine similarity of our method and ISIF~\cite{ye2019unsupervised} are compared in Fig.~\ref{F:visualization_distribution}. As can be seen, our method can obtain a more separable and compact distribution (most distances between positives are around 0) which indicates a more discriminative feature embedding can be learned by our model.

\textbf{The T-SNE visualization of learned embedding spaces on testing samples}: Our learned feature embedding from uncertainty momentum modeling not only performs well for training samples but also generalizes well to the unlabeled testing samples as demonstrated by Theorem.~\ref{T:1}. As shown by Fig.~\ref{F:visualization_tsne}, for the four evaluation datasets, the different-category testing samples are separable from each other as well as the same-category samples are closely clustered. These visualization results are consistent with the superior kNN classification performance of our method reported in Table.~\ref{T:cifar10+stl10} and Table.~\ref{T:cub200+product}.

\section{Conclusion}
Although existing unsupervised embedding learning methods have achieved promising results owing to their elaborate model trick designs and negative sample generation, the disregard for unlabeled sample outliers and global discrimination considerations severely limit their performance. In this paper, we thoroughly investigate the influence of unlabeled outliers in the given learning data and propose a novel unsupervised embedding learning network that specifically and explicitly models the sample uncertainty. The uncertainty is explored as momentum to facilitate unsupervised learning. To do so, each instance is represented by a multivariate Gaussian distribution of which the mean vector depicts the spatial localization of instance in the embedding space and the covariance matrix represents the feature uncertainty. In addition, multiple informative positive candidates for one instance can be readily drawn from the learned distributions which can enrich and balance the positive data. Our improvement of sample outliers is not only guaranteed by thorough theoretical analyses and justifications but also verified by extensive experiments on various evaluation benchmarks against the state-of-the-art unsupervised embedding learning approaches.

\ifCLASSOPTIONcaptionsoff
  \newpage
\fi

\bibliographystyle{IEEEtran}
\bibliography{egbib_udel_long}

%

\begin{IEEEbiography}[{\includegraphics[width=1in,height=1.25in,clip,keepaspectratio]{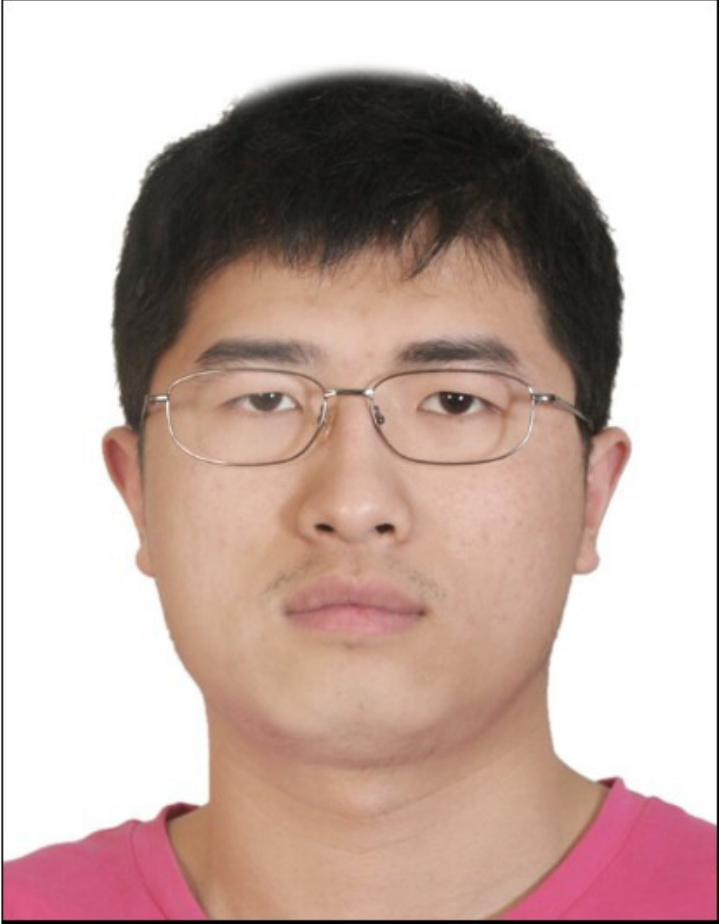}}]{Jiahuan Zhou}
	received his B.E. (2013) from Tsinghua University, the Ph.D. degree (2018) in the Department of Electrical Engineering \& Computer Science, Northwestern University (NU). During summer 2018, he was a research intern with Microsoft Research, Redmond, Washington. From 2019 to 2020, he was a Postdoctoral Fellow in Northwestern University. Currently, he is a Research Assistant Professor in Northwestern University. His current research interests include computer vision, pattern recognition, visual identification and machine learning. He has authored 15+ papers in top journals and conferences including IEEE T-PAMI, IEEE TIP, CVPR, ICCV, ECCV and so on. He serves as a regular reviewer member for a number of journals and conferences, e.g., T-PAMI, TIP, TCSVT, CVPR, ICCV, ECCV, NeurIPS, AAAI and so on.
\end{IEEEbiography}

\vskip -0.5in

\begin{IEEEbiography}[{\includegraphics[width=1in,height=1.25in,clip,keepaspectratio]{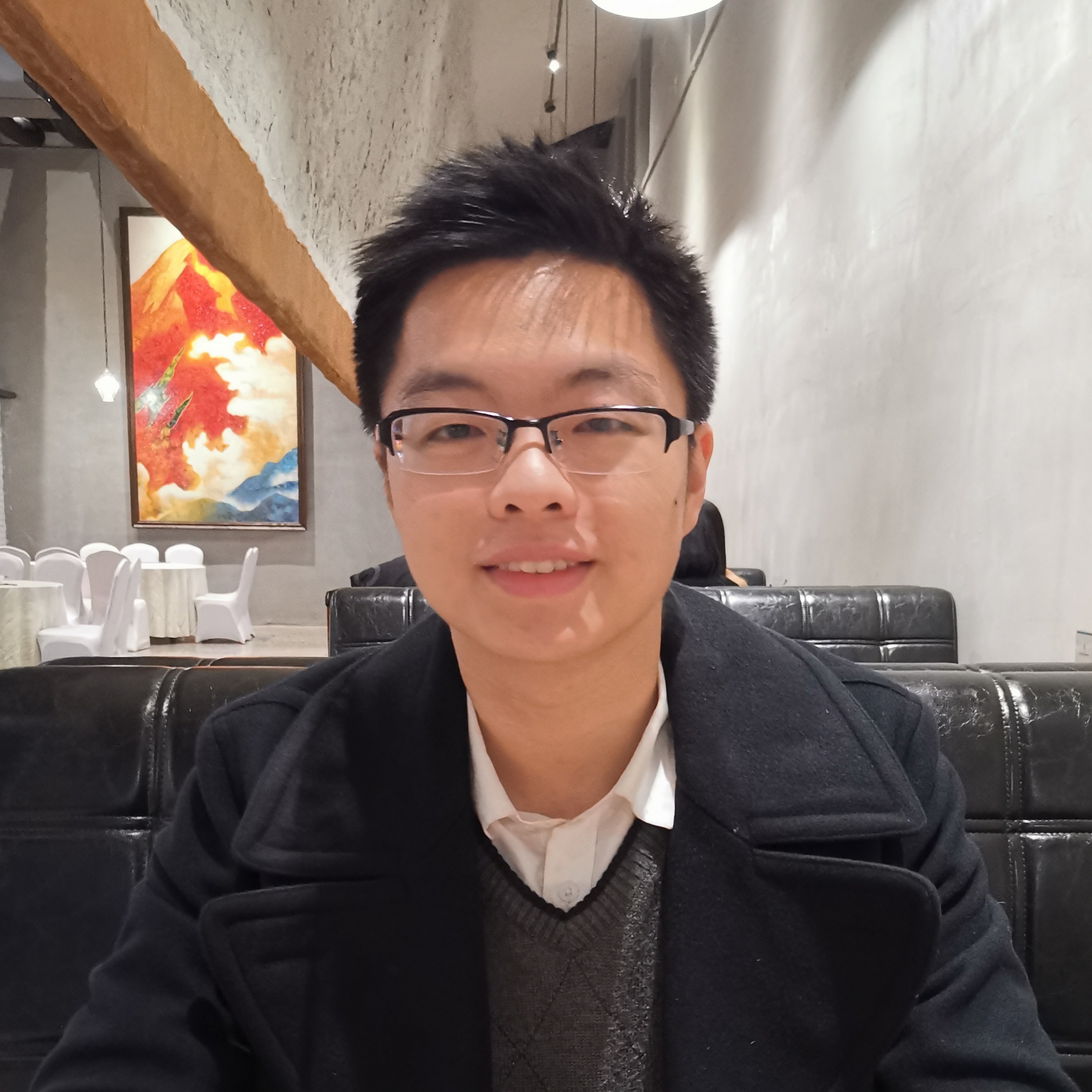}}]{Yansong Tang}
	is currently a research fellow in the Department of Engineering Science at the University of Oxford. His research interests lie in computer vision. Before coming to Oxford, he received Ph.D. degree and B.S. degree from Tsinghua University. He has also spent time at Visual Computing Group of Microsoft Research Asia (MSRA), and VCLA lab of University of California, Los Angeles (UCLA). He has authored 10 scientific papers in this area, where 6 papers are published in top journals and conferences including IEEE T-PAMI and CVPR. He serves as a regular reviewer member for a number of journals and conferences, e.g., T-PAMI, TIP, TCSVT, CVPR, ICCV, AAAI and so on. His Ph.D dissertation was awarded Excellent Doctoral Dissertation of Tsinghua University in 2020.
\end{IEEEbiography}

\vskip -0.5in

\begin{IEEEbiography}[{\includegraphics[width=1in,height=1.25in,clip,keepaspectratio]{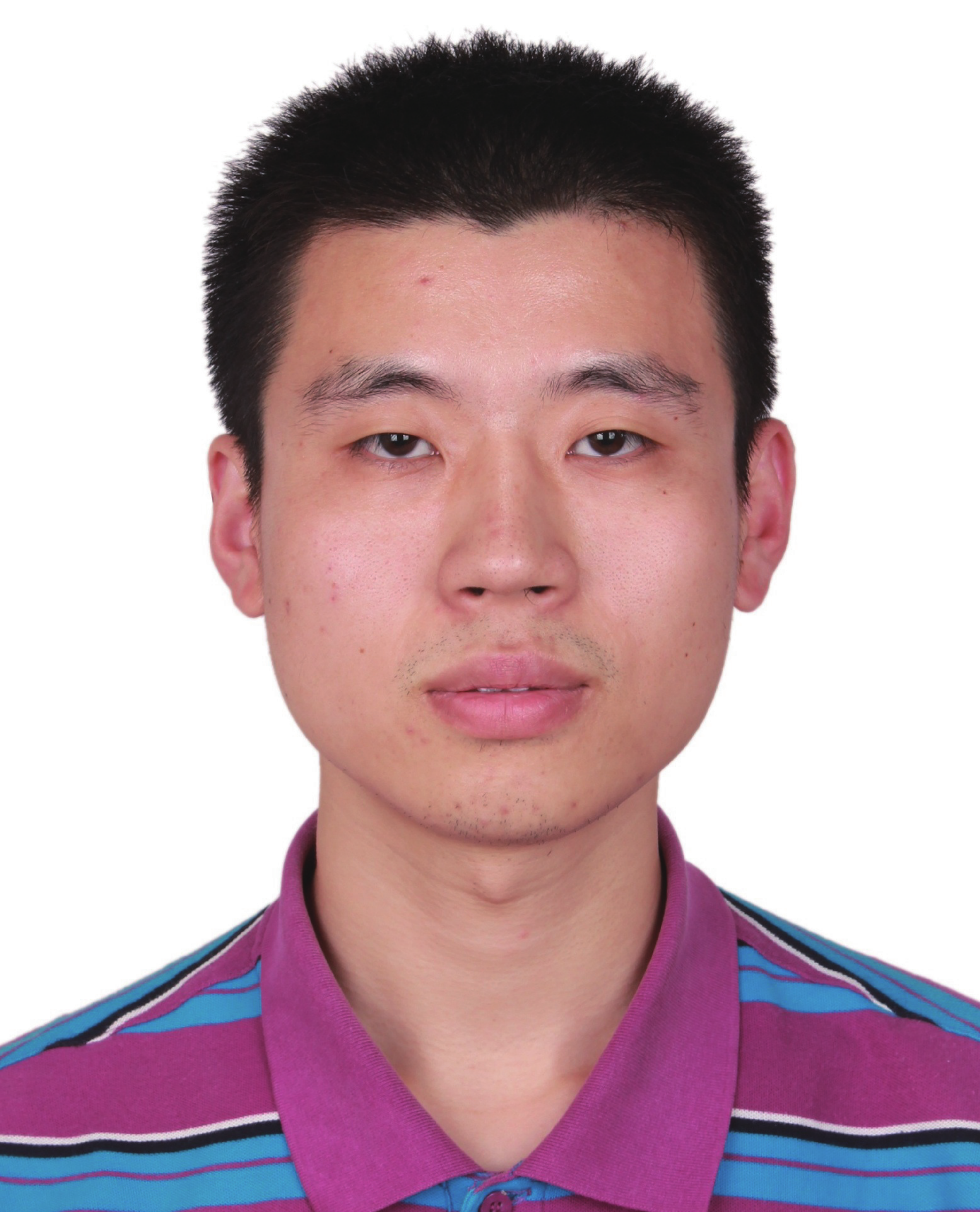}}]{Bing Su}
	received a B.S. degree in information engineering from Beijing Institute of Technology, Beijing, in 2010, and a Ph.D. degree in Electronic Engineering from Tsinghua University, Beijing, in 2016. From 2016 to 2018, he was an Assistant Professor of the Institute of Software, Chinese Academy of Sciences, Beijing. Currently, he is an Associate Professor in the Institute of Software, Chinese Academy of Sciences. His research interests include pattern recognition, computer vision, and machine learning.
\end{IEEEbiography}

\vskip -0.5in

\begin{IEEEbiography}[{\includegraphics[width=1in,height=1.25in,clip,keepaspectratio]{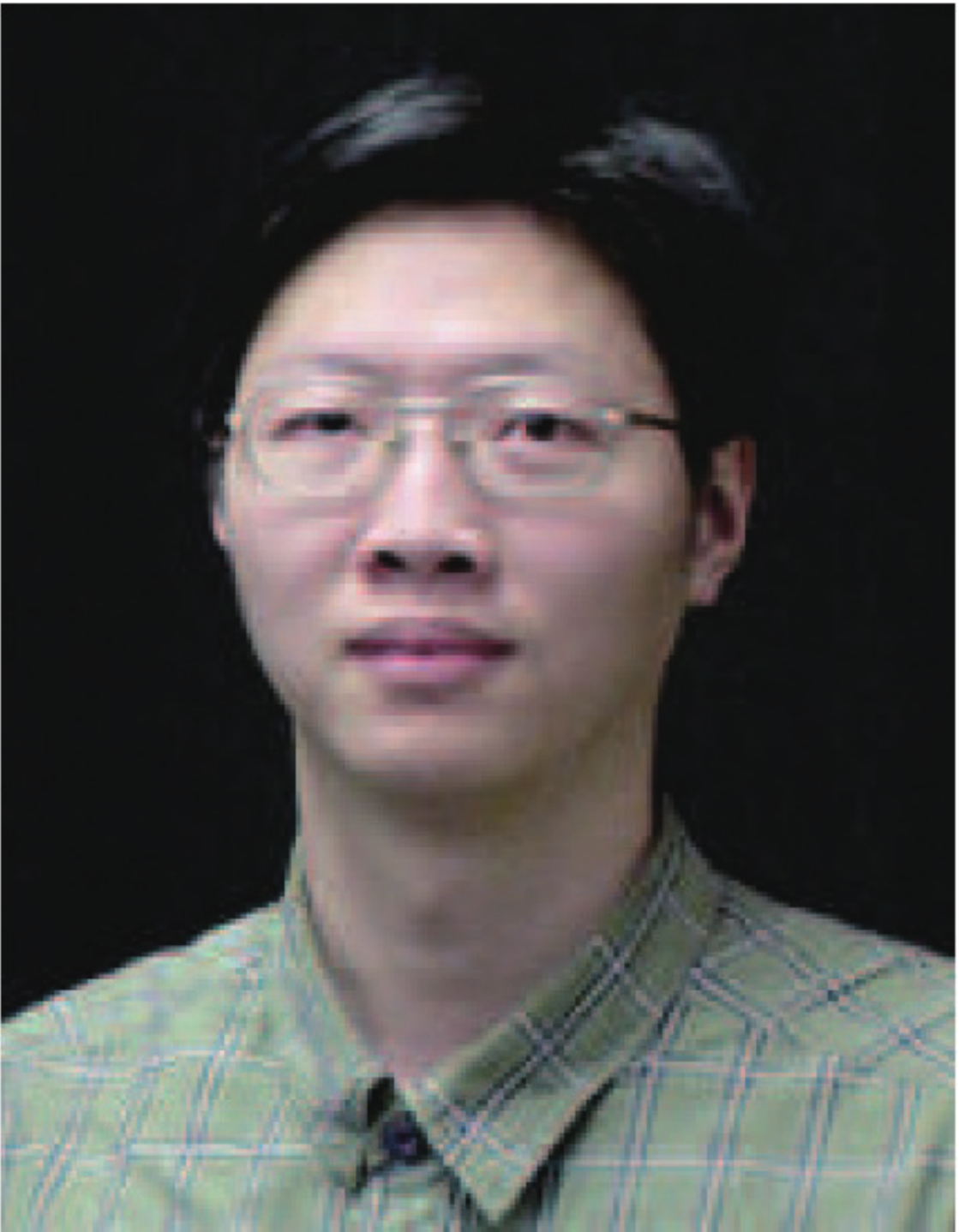}}]{Ying Wu}
	received the B.S. from Huazhong University of Science and Technology, Wuhan, China, in 1994, the M.S. from Tsinghua  University, Beijing, China, in 1997, and the Ph.D. in electrical and computer engineering from the University of Illinois at  Urbana-Champaign (UIUC), Urbana, Illinois, in 2001. From 1997 to 2001, he was a research assistant at the Beckman Institute for Advanced Science and Technology at UIUC. During  summer 1999 and 2000, he was a research intern with Microsoft Research, Redmond, Washington. In 2001, he joined the Department of Electrical and Computer Engineering at Northwestern University, Evanston, Illinois, as an assistant professor. He was  promoted to associate professor in 2007 and full professor in 2012. He is currently full professor of Electrical Engineering and  Computer Science at Northwestern University. His current research interests include computer vision, robotics, image and video  analysis, pattern recognition, machine learning, multimedia data mining, and human-computer interaction. He serves as the associate editor-in-chief for APR Journal of Machine Vision and Applications, and associate editors for IEEE Transactions on Pattern Analysis and Machine Intelligence, IEEE Transactions on Image Processing, IEEE Transactions on Circuits and Systems for Video Technology, SPIE Journal of Electronic Imaging. He served as Program Chair and Area Chairs for CVPR, ICCV and ECCV. He received the Robert T. Chien Award at UIUC in 2001, and the NSF CAREER award in 2003. He is a Fellow of the IEEE and the IAPR.
\end{IEEEbiography}

\end{document}